\documentclass{ecai} 

\usepackage{latexsym}
\usepackage{amssymb}
\usepackage{amsmath}
\usepackage{amsthm}
\usepackage{booktabs}
\usepackage{enumitem}
\usepackage{graphicx}
\usepackage{xcolor}
\usepackage{svg}
\usepackage{hyperref}
\usepackage{caption}

\usepackage{pgfplots}
\pgfplotsset{compat = newest}
\definecolor{color1}{RGB}{0,114,189}
\definecolor{color2}{RGB}{217,83,25}
\definecolor{color3}{RGB}{237,177,32}
\definecolor{color4}{RGB}{126,47,142}
\definecolor{color5}{RGB}{119,172,48}
\definecolor{color6}{RGB}{77,190,238}
\definecolor{color7}{RGB}{10,10,10}
\definecolor{color8}{RGB}{247,129,191}

\newcommand{\BibTeX}{B\kern-.05em{\sc i\kern-.025em b}\kern-.08em\TeX}

\begin{document}


\begin{frontmatter}


\paperid{123} 





\title{Visual RAG: Expanding MLLM visual knowledge without fine-tuning}




\author[A]{\fnms{Mirco}~\snm{Bonomo}}
\author[A]{\fnms{Simone}~\snm{Bianco}\thanks{Corresponding Author. Email: simone.bianco@unimib.it}
}

\address[A]{University of Milano-Bicocca, Italy}


\begin{abstract}
Multimodal Large Language Models (MLLMs) have achieved notable performance in computer vision tasks that require reasoning across visual and textual modalities, yet their capabilities are limited to their pre-trained data, requiring extensive fine-tuning for updates.
Recent researches have explored the use of In-Context Learning (ICL) to overcome these challenges by providing a set of demonstrating examples as context to augment MLLMs performance in several tasks, showing that many-shot ICL leads to substantial improvements compared to few-shot ICL. However, the reliance on numerous demonstrating examples and the limited MLLMs context windows presents significant obstacles, severely restricting the scope of applications for these models. 

This paper aims to address these challenges by introducing a novel approach, Visual RAG, that synergically combines the MLLMs capability to learn from the context,  with a retrieval mechanism.
The crux of this approach is to ensure to augment the MLLM knowledge by selecting only the most relevant demonstrating examples for the query, pushing it to learn by analogy.
In this way, relying on the new information provided dynamically during inference time, the resulting system is not limited to the knowledge extracted from the training data, but can be updated rapidly and easily without fine-tuning. Furthermore, this greatly reduces the computational costs for improving the model image classification performance, and augments the model knowledge to new visual domains and tasks it was not trained for.
Extensive experiments are carried out on eight different datasets in the state of the art spanning several domains and image classification tasks. The experimental results show that the proposed Visual RAG, compared to the most recent state of the art (i.e., many-shot ICL), is able to obtain an accuracy that is very close or even higher (approx. +2\% improvement on average) while using a much smaller set of demonstrating examples (approx. only 23\% on average).

\end{abstract}

\end{frontmatter}


\section{Introduction}

The rapid advancement of Large Language Models (LLMs) \cite{zhao2023survey} has unlocked extraordinary capabilities \cite{wei2022emergent}, making them valuable tools in numerous applications \cite{kaddour2023challenges}. However, this technological leap has not been without its challenges, such as the issue of outdated knowledge \cite{kaddour2023challenges}. Although fine-tuning was initially seen as a solution to this problem, the process is complex and costly.

This has encouraged research into more efficient methods to update or expand the knowledge base of these models. In this context, Retrieval-Augmented Generation (RAG) has emerged as a promising solution \cite{lewis2020retrieval}: by leveraging LLMs' ability to process information provided in the context (ICL) \cite{dong2024surveyincontextlearning}, RAG can augment the knowledge possessed by these models. The innovative aspect of RAG lies in its selective approach to provide information to the LLM during queries: RAG furnishes the model with only the most relevant document chunks, carefully selected from an external knowledge base based on semantic similarity between the query and these chunks.

This approach has gained significant traction in the field of natural language processing, enabling continuous knowledge updates with minimal effort and cost.
With the introduction of Multimodal Large Language Models (MLLMs) \cite{Yin_2024,zhang2024mm}, models have acquired the ability to work on multimodal content, including, among others, images.
This has opened up the possibility of exploiting MLLMs in the field of computer vision, unleashing the potential for a whole new range of applications. 
Existing studies \cite{han2024doesgpt4visionadaptdistribution,zhang2024outofdistributiongeneralizationmultimodallarge} have only begun to scratch the surface of these models' ability to exploit ICL with multimodal content, such as images. 

Taking inspiration from the RAG solutions currently used for textual information, this work aims to 
develop a solution that expands the visual knowledge of an MLLM beyond its training without requiring a fine-tuning of the model, but allowing an easy and quick update of knowledge with reduced costs.

The main contributions of this work can be summarized as follows:
\begin{itemize}
    \item[-] We introduce Visual RAG, a novel framework that combines retrieval-augmented generation with multimodal large language models (MLLMs) to dynamically expand their visual knowledge without requiring fine-tuning. 
    \item[-] Our method employs a robust embedding model and a retrieval mechanism to extract the most relevant examples from an image-based knowledge base, optimizing task-specific context for improved performance.
    \item[-] Our approach seamlessly integrates visual retrieval with MLLM in-context learning capabilities, enabling dynamic adaptation to new tasks and domains exhibiting strong generalization across eight diverse datasets. 
    \item[-] The proposed approach achieves comparable or superior performance to the state-of-the-art many-shot ICL method with significantly fewer demonstrating examples (about 23\% on average), reducing computational costs while maintaining scalability.
\end{itemize}


\section{Related Work}

{\bf{MLLM models}} The term MLLM refers to those models that combine the NLP capabilities possessed by classic LLMs with the ability to receive, reason, and output with multimodal information  \citep{wu2023multimodal,Yin_2024}. 
While our focus primarily lies on visual modalities, it is important to note that MLLMs can also incorporate other modalities such as videos and audios.
Recent state-of-the-art models, including LLaVA \citep{liu2024improved}, GPT-4o \citep{openai2024gpt4technicalreport} and Gemini \citep{geminiteam2024gemini15unlockingmultimodal}, exemplify the potential of MLLMs.
Characterized by their massive scale, these models are accessible through web APIs. 
However, despite their rapid development and impressive performance, the full capabilities of MLLMs remain untapped.
The primary challenges the research community is facing include the limited context windows of these models, which restrict the amount of information that can be processed concurrently.  
Additionally, the reliance only on training data can lead to outdated knowledge and poses challenges in keeping these models up-to-date since fine-tuning, although effective, is a computationally expensive process \citep{wang2024knowledge}. Furthermore, fine-tuning can also increase the model's tendency to hallucinate \citep{gekhman2024does}.

{\bf{Multimodal ICL}}
In the case of text-based In-Context Learning (ICL)\citep{dong2024surveyincontextlearning}, LLMs demonstrate remarkable ability to learn from a few textual information provided in the context. This enables them to tackle complex unseen tasks in a few-shot manner \citep{brown2020language}. The strength of ICL lies in its training-free nature, allowing a flexible integration into various frameworks and applications \citep{Yin_2024}. 
Multimodal In-Context Learning (M-ICL) extends this concept to incorporate multiple modalities like images, videos and audios. In essence, MLLMs learn from a few task-specific examples and generalize to novel, yet similar, questions \citep{Yin_2024,zhang2024outofdistributiongeneralizationmultimodallarge}.
However, LLMs, including multimodal ones, face a significant limitation: the context window size, as witnessed by the design of  specific solutions to extend it (e.g., \cite{chen2023extending}). This constraint, measured in tokens, restricts the size of the input a model can process during generation. Although advances in research have led to larger context windows, enabling more examples to be included in prompts, the associated costs escalate. 
Furthermore, even with extremely large context windows, the so-called 'lost in the middle' effect, as observed in \citep{liu2023lostmiddlelanguagemodels}, remains a significant concern: models often struggle to effectively utilize all the information within very large contexts, tending to rely more heavily on data at the beginning and at the end of the input.
This underscores the critical importance of efficient ICL techniques.

{\bf{Textual RAG}}
Retrieval-Augmented Generation (RAG) \cite{lewis2020retrieval} is an increasingly prominent approach in the field of LLMs for several reasons, including one of particular interest to this research: enabling these models to utilize in the response generation process new textual information that was not provided during training.
RAG synergistically merges LLMs’ intrinsic knowledge with the vast, dynamic repositories of external texts (e.g., documents, papers, websites) \citep{gao2024retrievalaugmentedgenerationlargelanguage,izacard2022atlasfewshotlearningretrieval}.
This allows the use of a pre-trained model in knowledge-intensive tasks, continuously updating its knowledge and integrating domain-specific information without requiring a fine-tuning process and all the associated challenges.
The diffusion of this approach is rapidly increasing, making it a key technology for the use of LLMs in real-world applications. 
A classic RAG solution, such as the one used for question-answering problems, primarily leverages three steps \citep{izacard2022atlasfewshotlearningretrieval}:
\begin{itemize}
  \item[-] Indexing: Documents containing the information are divided into chunks, transformed into embeddings, and stored indexed in a vector Data Base (DB).
  \item[-] Retrieval: The query (question) is transformed into an embedding using the same approach as indexing, and the most semantically similar chunks are retrieved.
  \item[-] Generation: The LLM, here called "generator", is queried with the original request (question), providing the retrieved text chunks in the context, and the model is asked to use this information to help answering the question.
\end{itemize}




\section{The proposed Visual RAG}

To investigate the effectiveness of a RAG system in the field of computer vision, we implemented a solution that allows to expand the vision capabilities of a MLLM by using a Knowledge Base composed by example images.
We implement a retriever capable of working on images instead of text chunks, appropriately combined with one of the best available MLLM, i.e., Gemini  \citep{geminiteam2024gemini15unlockingmultimodal}, to leverage its ICL capabilities.  
Inspired by \citep{yixing2024manyshot} we assess the performance of the proposed solution by querying it using several computer vision datasets that serve as benchmarks.

\begin{figure*}
\centering
\includegraphics[trim={1.cm -0.5cm 0.7cm 0},clip,width=\textwidth]{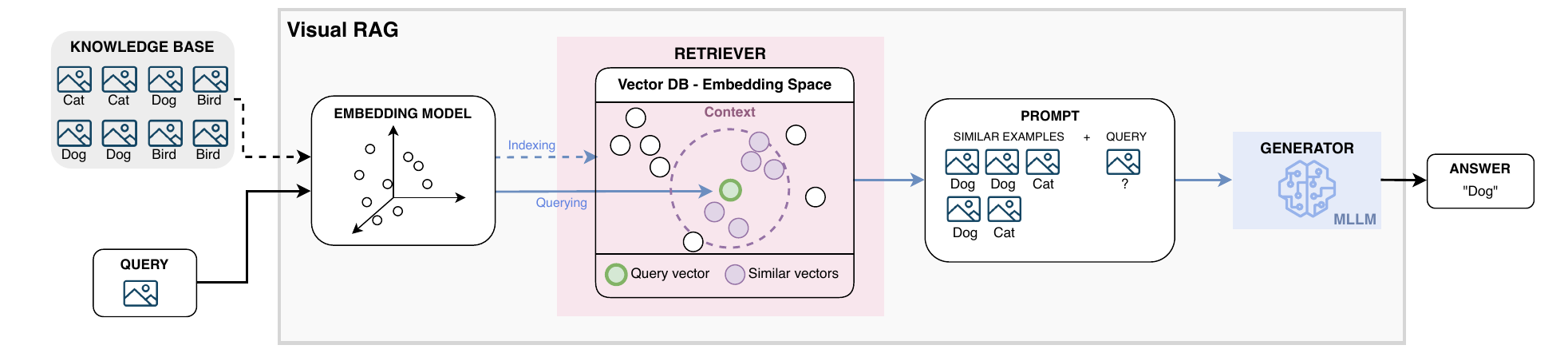}
\caption{Schematic representation of the architecture of Visual RAG: the query image and the Knowledge Base (KB) constituted by image-label pairs are fed to the embedding model to obtain their vector representations. In the embedding spaces the retriever returns the requested number of demonstrating examples by selecting the images in the KB whose vectors are closer to the vector of the query image. The generator (i.e., MLLM) is then prompted with the query image and the demonstrating examples to predict the label for the query image.\\}
\label{fig:vRAGarchitecture}
\end{figure*}

RAG systems are employed in the field of LLMs to incorporate external textual knowledge  into the reasoning processes of models by leveraging ICL \citep{gao2024retrievalaugmentedgenerationlargelanguage}.
Our idea is to apply this methodology to the field of computer vision, allowing MLLM models to integrate external knowledge for computer vision tasks such as image classification (and beyond).
To achieve this, we replace the traditional text-based knowledge base with a new one containing images labeled with their respective class/classes.
This knowledge base is then converted into embeddings capable of capturing the visual information required for the classification task and subsequently indexed using a vector database.
The implemented retriever operates similarly to a retriever in a conventional RAG system. However, instead of searching for text chunks that are semantically related to the input query, it retrieves images that are most similar to the image to be classified. 
Finally, these images are provided to the generator as examples to be used as a reference to classify new instances.
Figure~\ref{fig:vRAGarchitecture} shows the architecture of the proposed Visual RAG.

\subsection{Indexing}
For each dataset considered, the Knowledge Base (KB) is composed of the images contained in its corresponding demo set.
To facilitate and optimize the subsequent retrieval phase, the KB must first be indexed. 
First, the images are encoded into a vector representation using an embedding model and then stored in a vector DB. 
Given the nature of the task, we selected CLIP \citep{radford2021learningtransferablevisualmodels}, one of the state-of-the-art models to represent visual and text information. It was chosen not only because it is a multimodal model capable of processing images but also because of its ability to generalize, a crucial characteristic for developing a visual RAG solution capable of performing effectively across diverse contexts.
Consequently, FAISS \citep{douze2024faisslibrary} was selected as the underlying vector DB/retriever for our solution. 
Its efficiency and lightweight design allow it to operate effectively on standard hardware, delivering rapid response times even when handling massive datasets. Additionally, it prepares the solution to scale efficiently in case larger datasets will be used in the future.

We create the indexes using the IndexFlatL2 method provided by FAISS \citep{douze2024faisslibrary}, a basic implementation that does not rely on complex index structures but stores the vectors without compression as a flat array. Although this approach is not the fastest one, it is highly accurate, making it the most suitable choice for small datasets like ours.

\subsection{Retrieval}
As mentioned in the previous section, we use FAISS to index the Knowledge Base in IndexFlatL2 mode.
This configuration employs a brute-force search method across the entire index, making it highly accurate but less efficient. Specifically, it computes the similarity between the input image and each image in the knowledge base by measuring the L2 distance (i.e., Euclidean distance) between their respective vectors and retrieves the desired number of examples that are closest (nearest neighbors) according to this metric. \\
Given two $n$-dimensional vectors $\mathbf{x}=\left[x_1,\ldots,x_n\right]$ and $\mathbf{y}=\left[y_1,\ldots,y_n\right]$, the L2 distance between them is defined as:

\begin{equation}
d(\mathbf{x}, \mathbf{y}) = \sqrt{\sum_{i=1}^{n} (x_i - y_i)^2}    
\label{eq:l2_distance}
\end{equation}


\subsection{Generator}
The visual RAG solution requires a model with remarkable ICL capabilities and the possibility to submit several images in the request.
Furthermore, since ICL capabilities are most effectively exhibited by larger models \citep{wei2022emergentabilitieslargelanguage} that cannot be utilized with common hardware, the focus shifts to the larger models accessible with APIs.
These requirements narrow the research to two options: ChatGPT and Gemini. 
We chose to conduct all the experiments using Gemini because, as observed in  \citep{yixing2024manyshot}, Gemini 1.5 Pro demonstrated a more linear improvement across different scenarios as the number of demonstrating examples increases, leading to a more stable model.
For a fair comparison, we use the same model version used in \citep{yixing2024manyshot}: the “gemini-1.5-pro-preview-0409” model. \\

To leverage the model's ICL capabilities, the RAG Solution employs the prompt detailed in Appendix A \citep{yixing2024manyshot}. The retrieved examples are presented as enhanced context to the generator in the form of <image, label> pairs, serving as demonstrations. The input image from the test set, which requires classification, is then included in the prompt using the same structure but without its corresponding label. This prompts the model to predict the appropriate class.

Additionally, the prompt explicitly defines the expected response format, ensuring that the model output adheres to a structure that simplifies the extraction of the predicted class from the generated text.

\section{Experimental setup}

\subsection{Datasets}
\begin{figure}
\centering
\resizebox{0.95\columnwidth}{!}{
    \begin{tabular}{ccccc}
         \includegraphics[width=0.5\textwidth,height=0.5\textwidth]{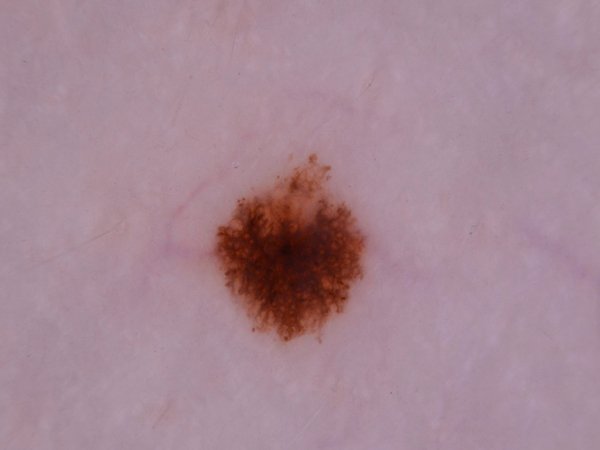} &
         \includegraphics[width=0.5\textwidth,height=0.5\textwidth]{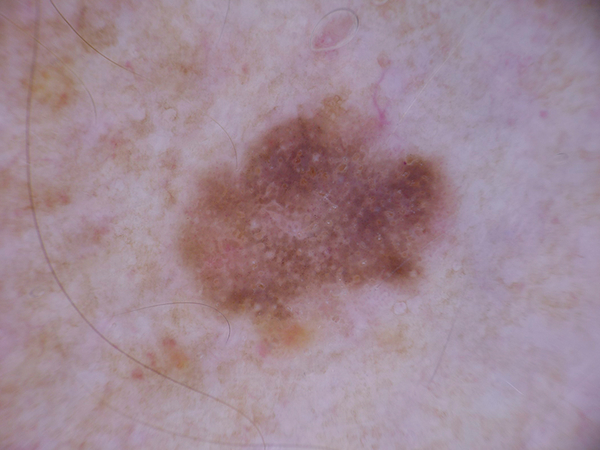} &
         \includegraphics[width=0.5\textwidth,height=0.5\textwidth]{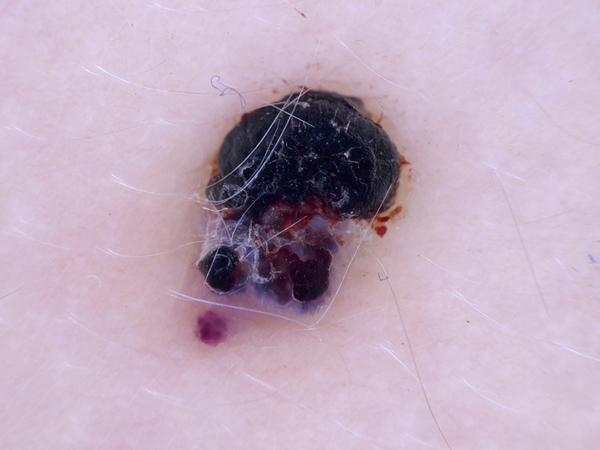} &
         \includegraphics[width=0.5\textwidth,height=0.5\textwidth]{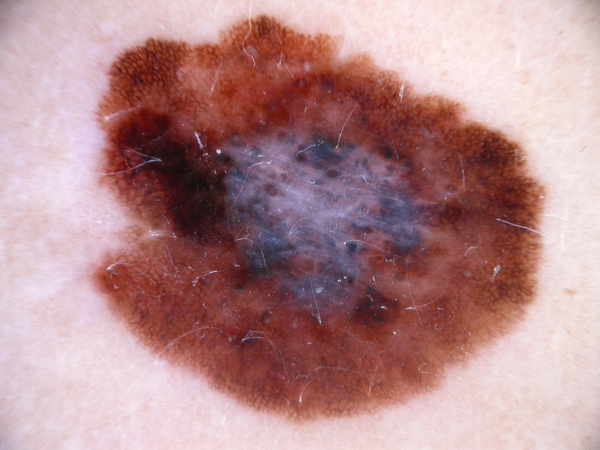} &
         \includegraphics[width=0.5\textwidth,height=0.5\textwidth]{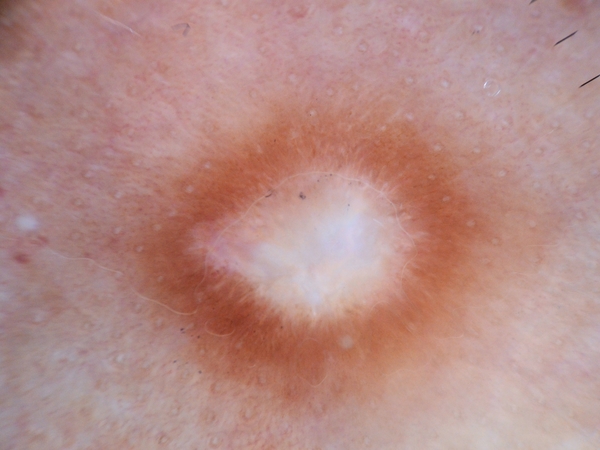} \\
         \includegraphics[width=0.5\textwidth,height=0.5\textwidth]{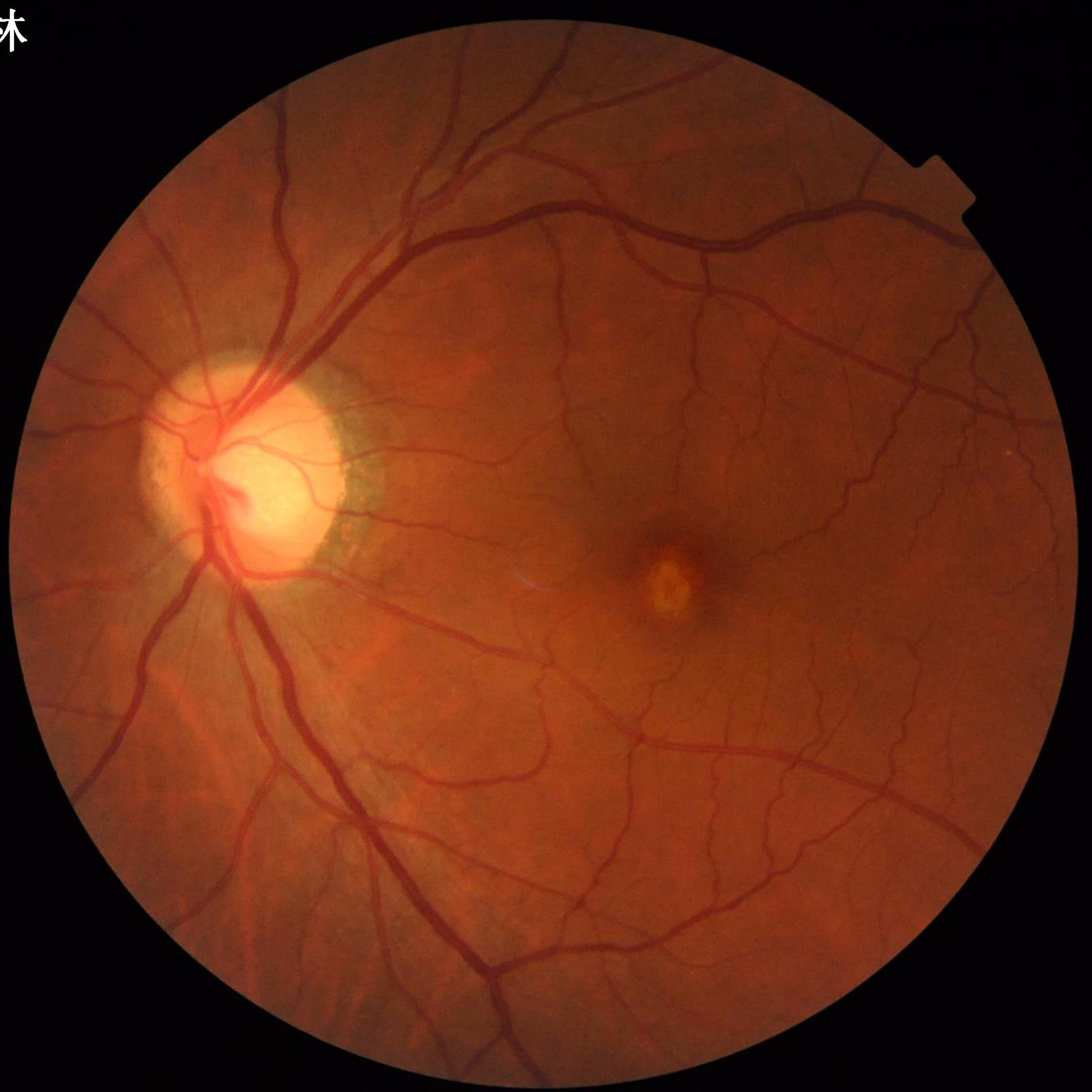} &
         \includegraphics[width=0.5\textwidth,height=0.5\textwidth]{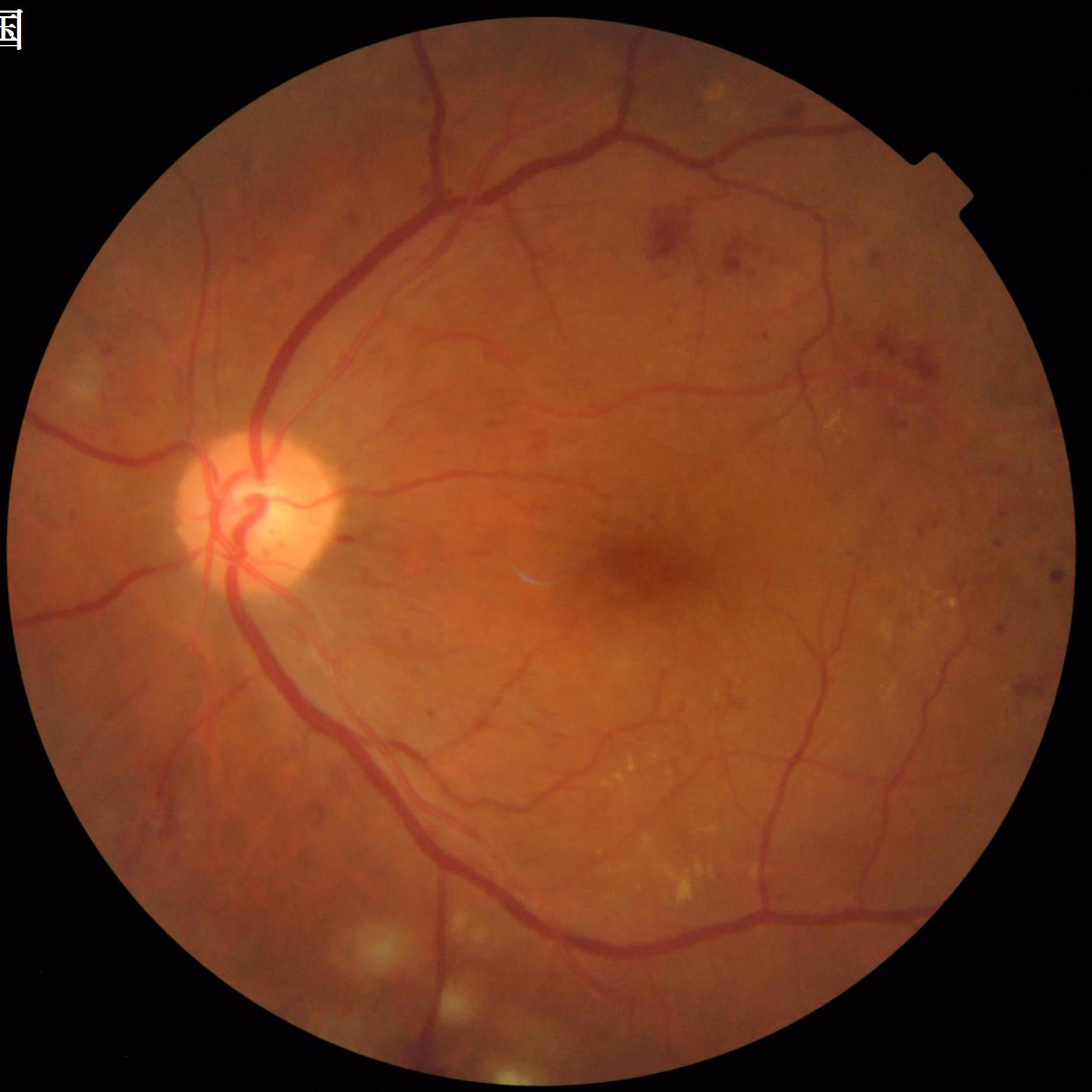} &
         \includegraphics[width=0.5\textwidth,height=0.5\textwidth]{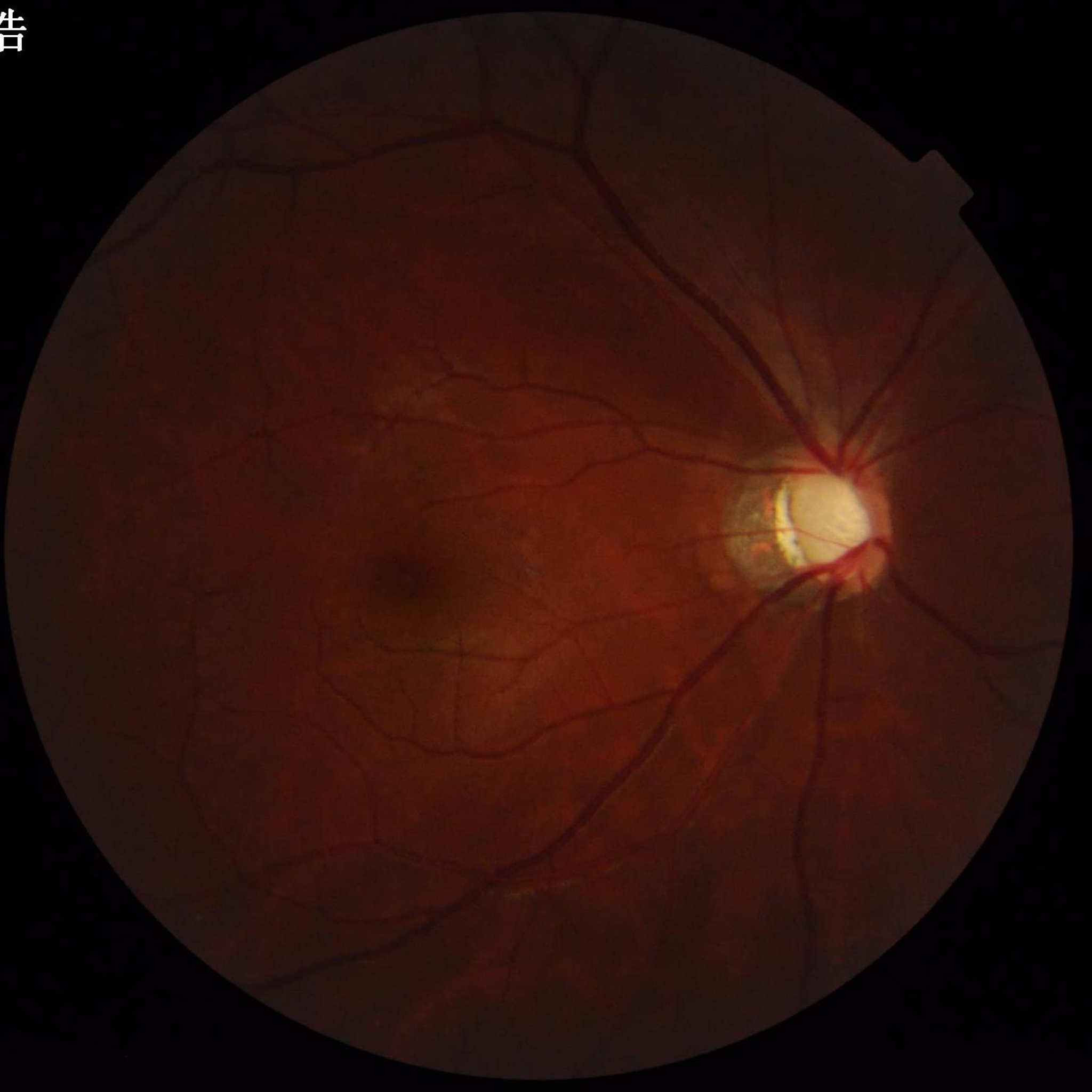} &
         \includegraphics[width=0.5\textwidth,height=0.5\textwidth]{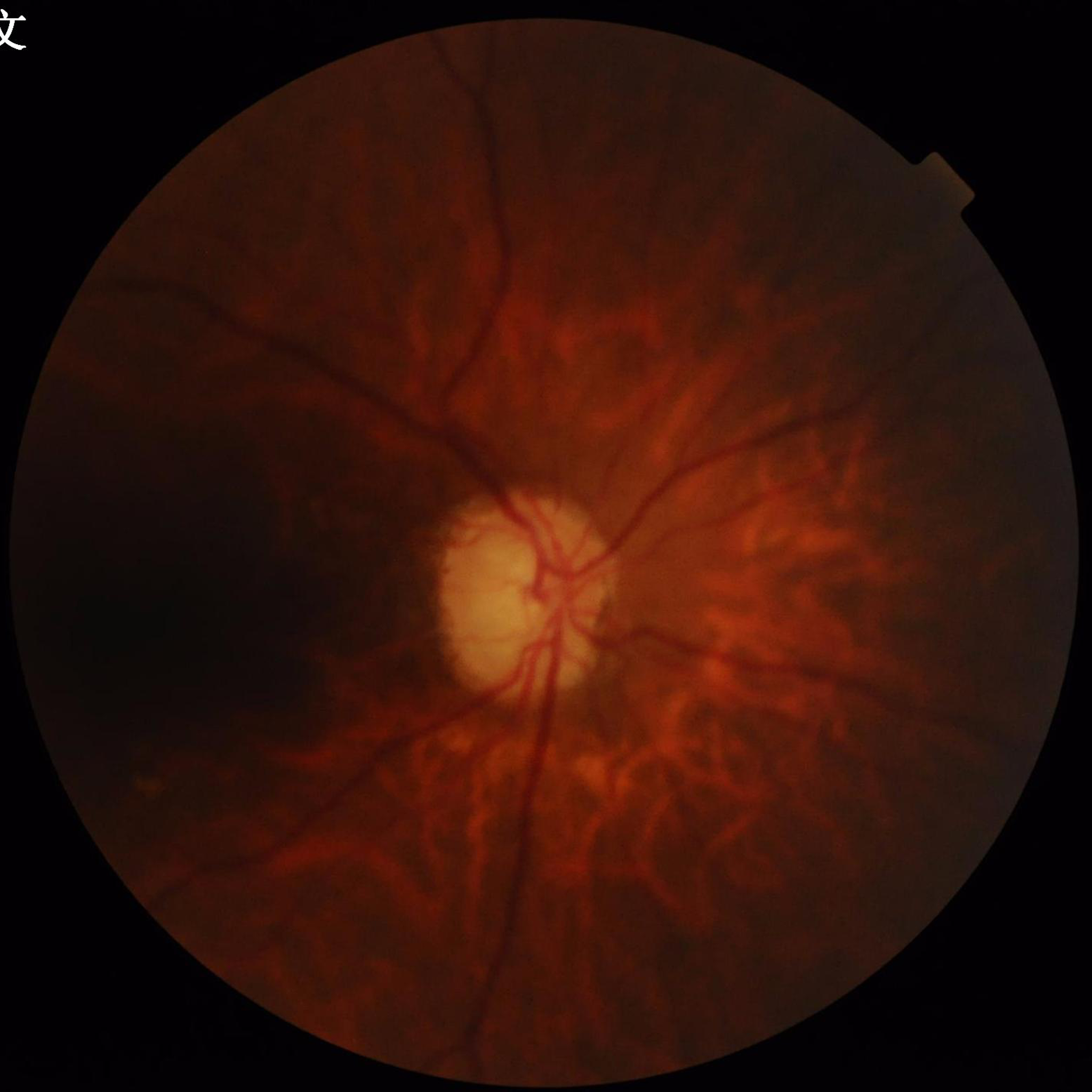} &
         \includegraphics[width=0.5\textwidth,height=0.5\textwidth]{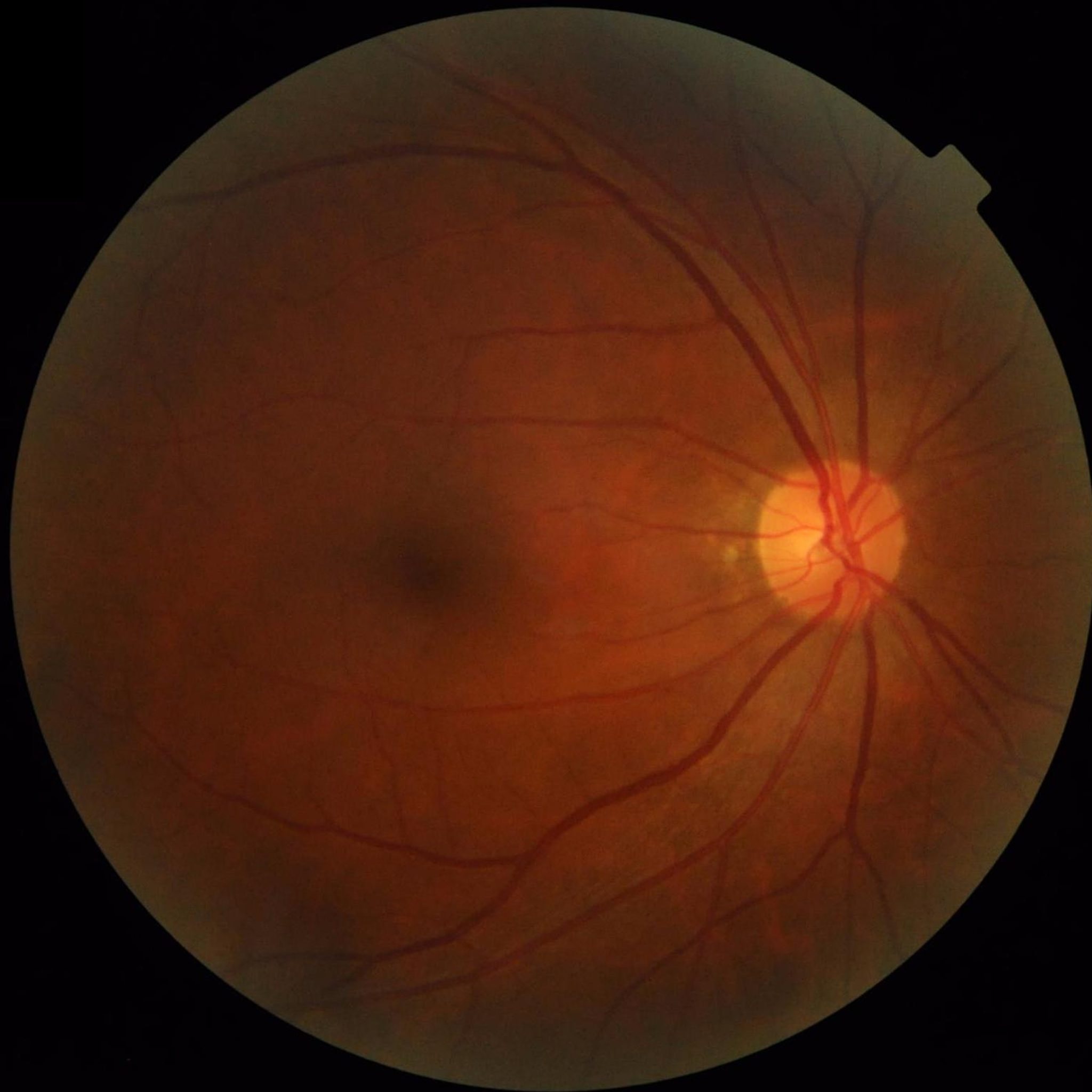} \\
         \includegraphics[width=0.5\textwidth,height=0.5\textwidth]{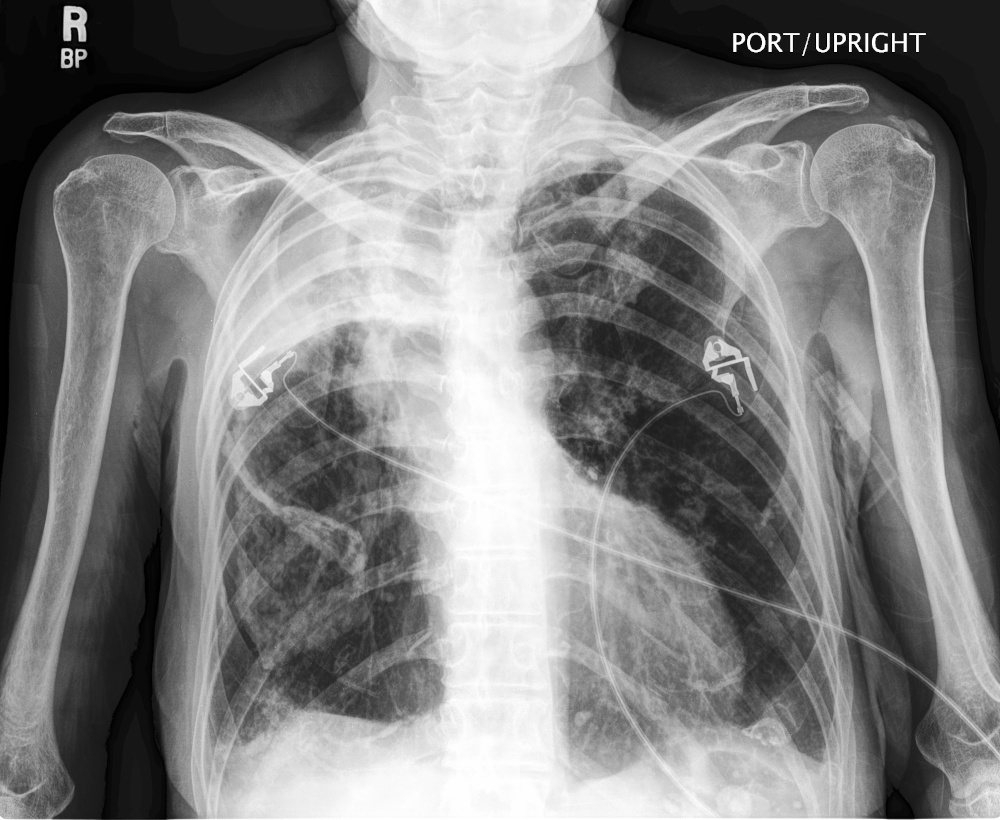} &
         \includegraphics[width=0.5\textwidth,height=0.5\textwidth]{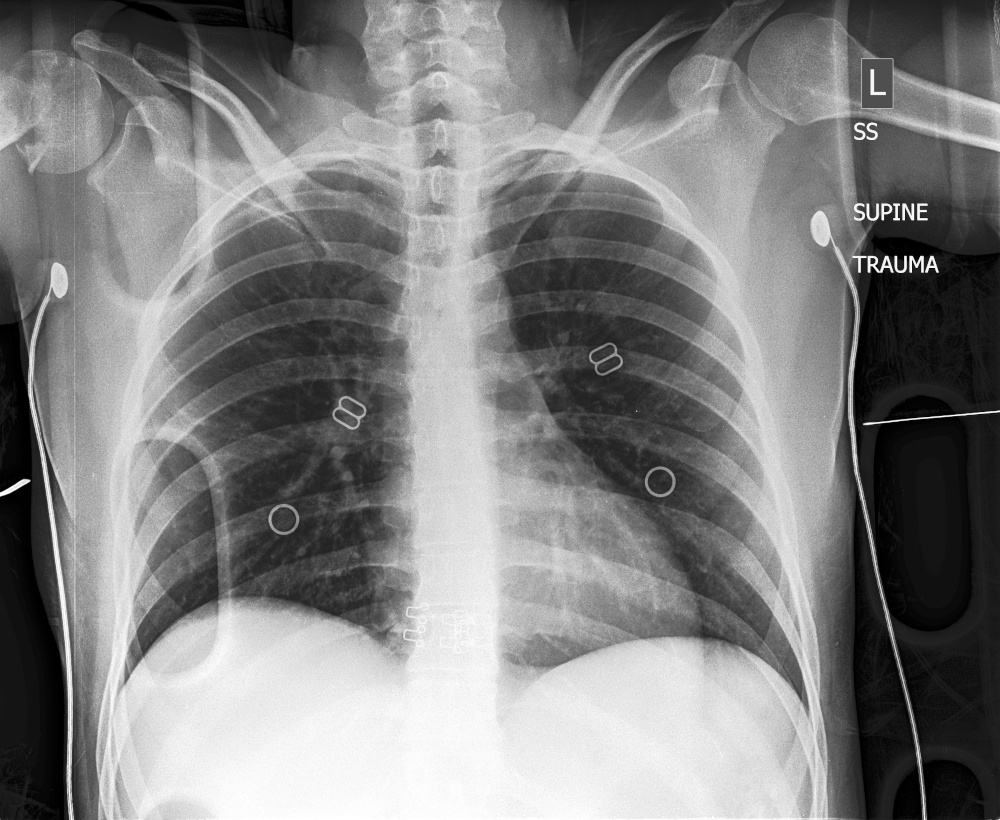} &
         \includegraphics[width=0.5\textwidth,height=0.5\textwidth]{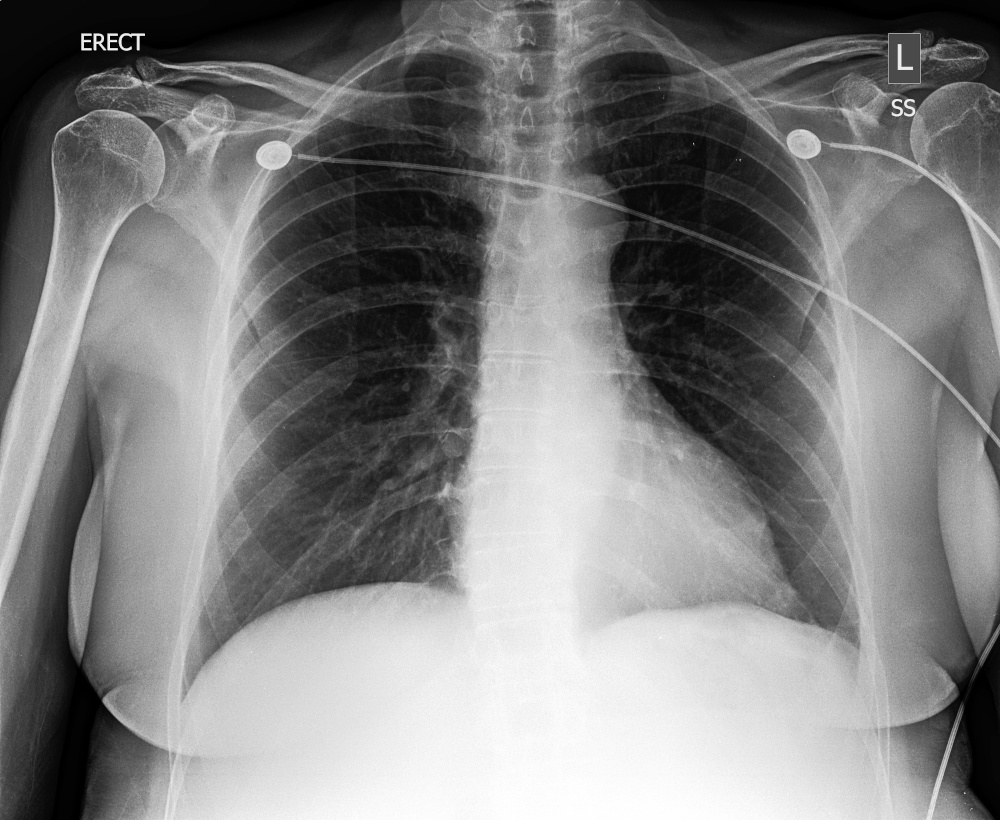} &
         \includegraphics[width=0.5\textwidth,height=0.5\textwidth]{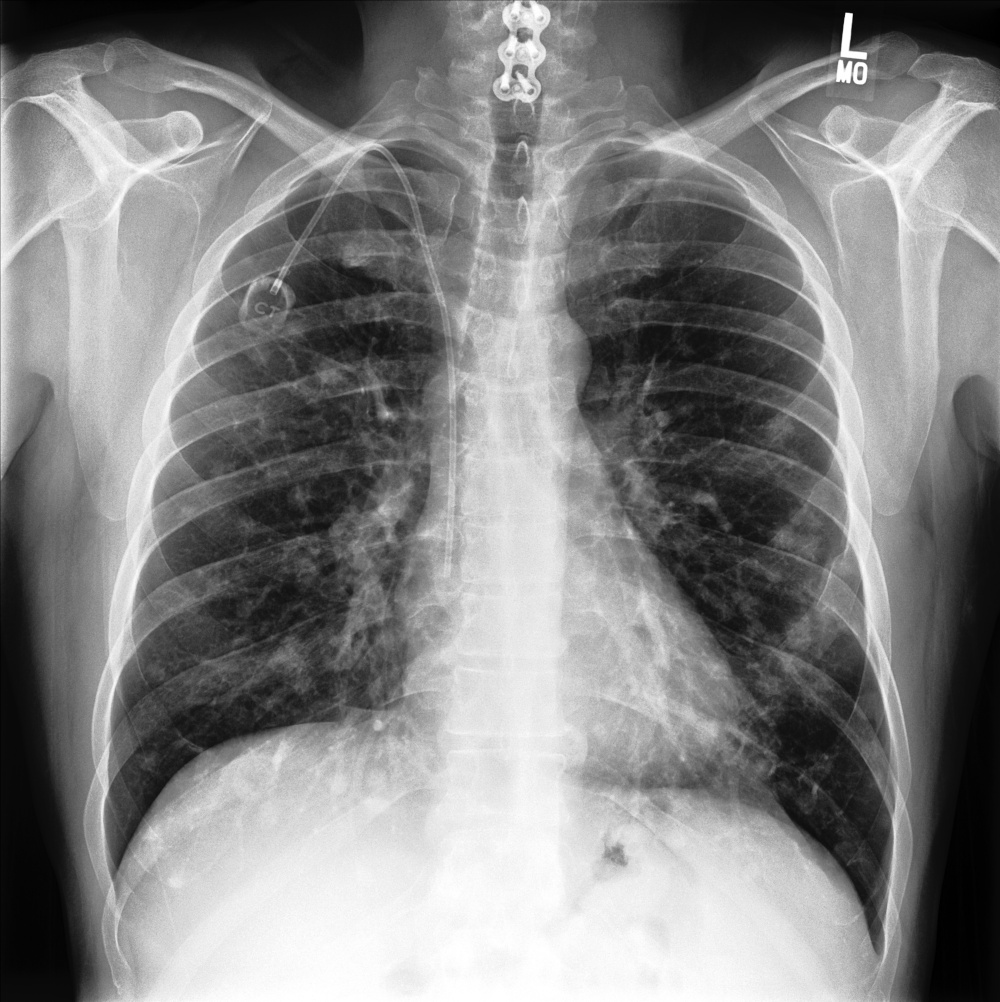} &
         \includegraphics[width=0.5\textwidth,height=0.5\textwidth]{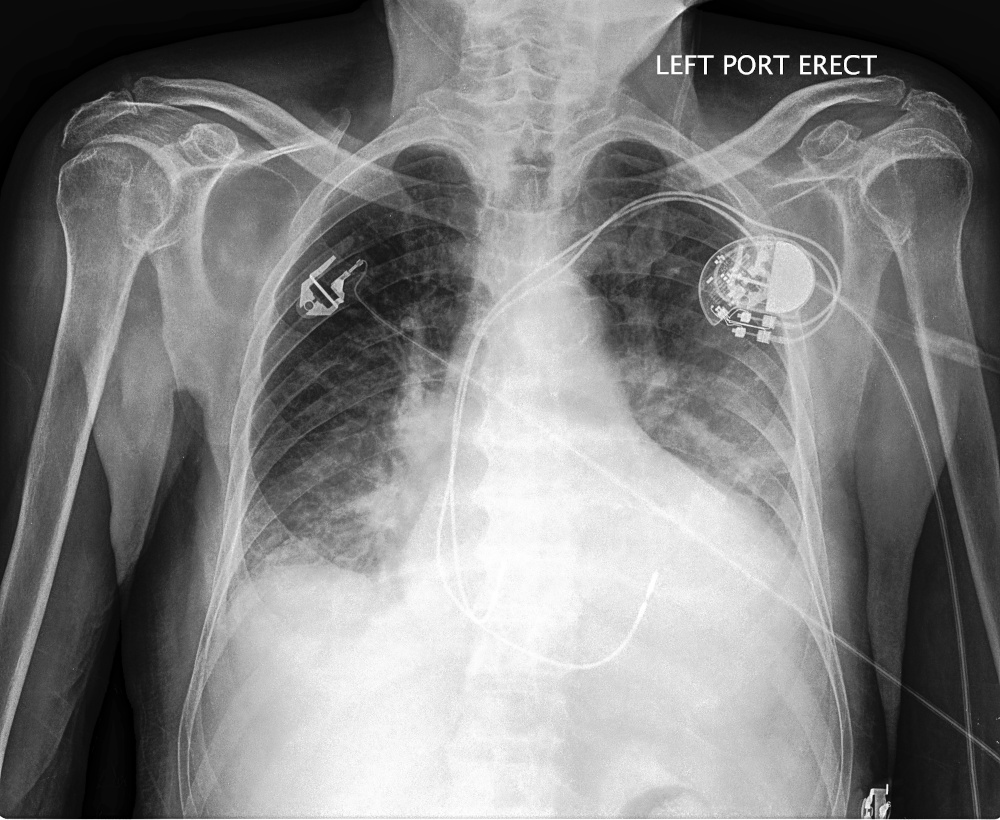} \\
         \includegraphics[width=0.5\textwidth,height=0.5\textwidth]{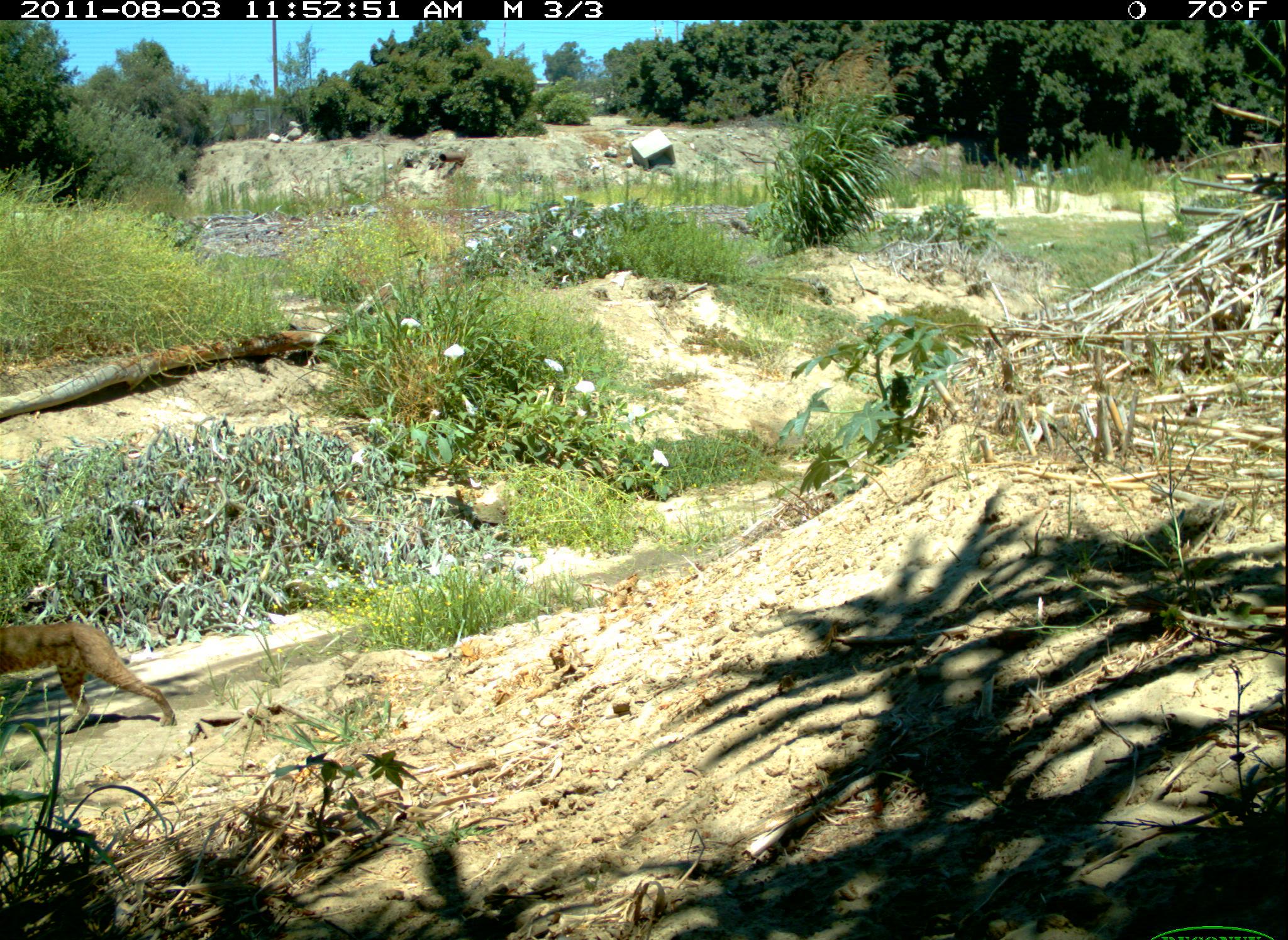} &
         \includegraphics[width=0.5\textwidth,height=0.5\textwidth]{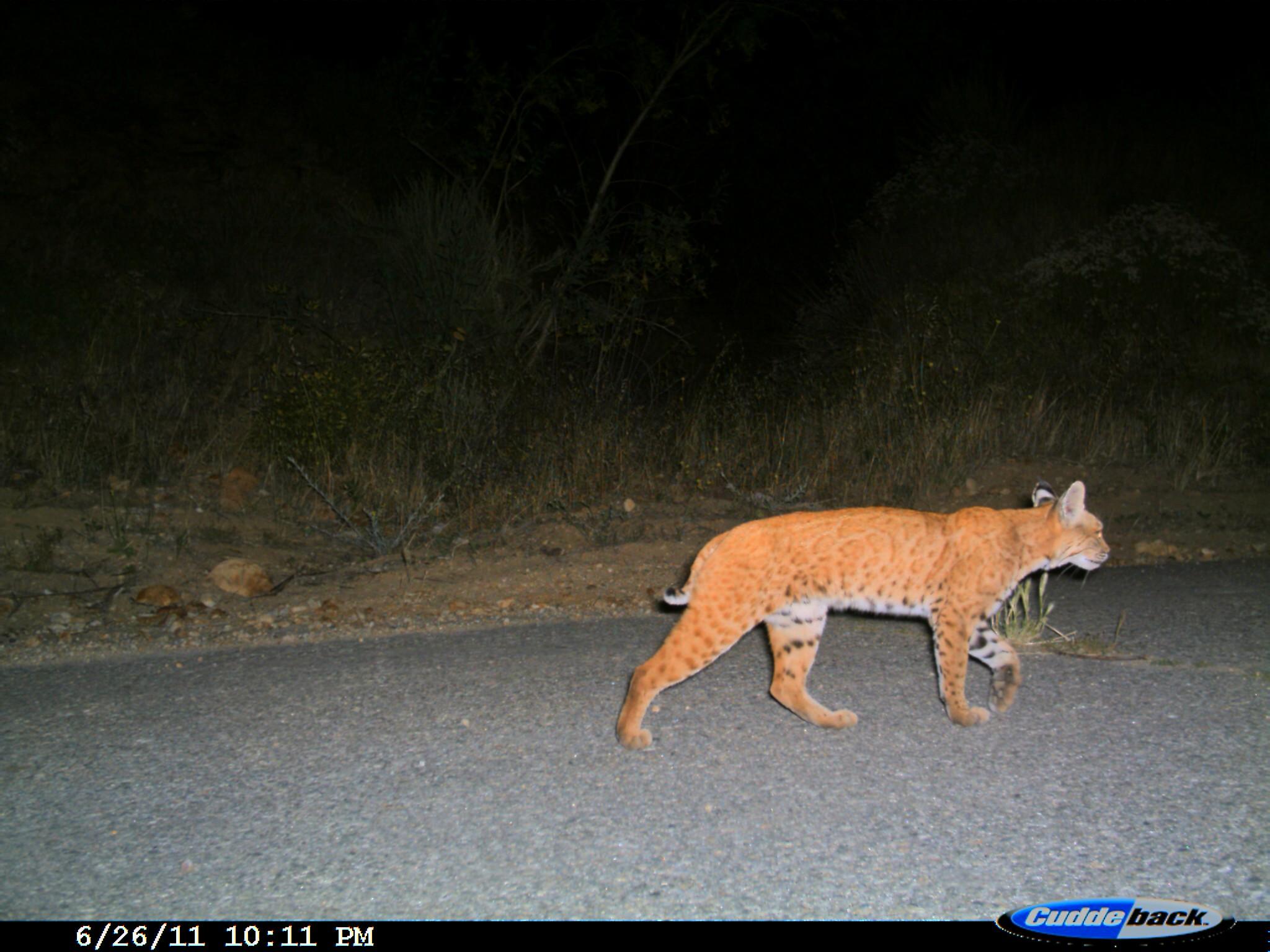} &
         \includegraphics[width=0.5\textwidth,height=0.5\textwidth]{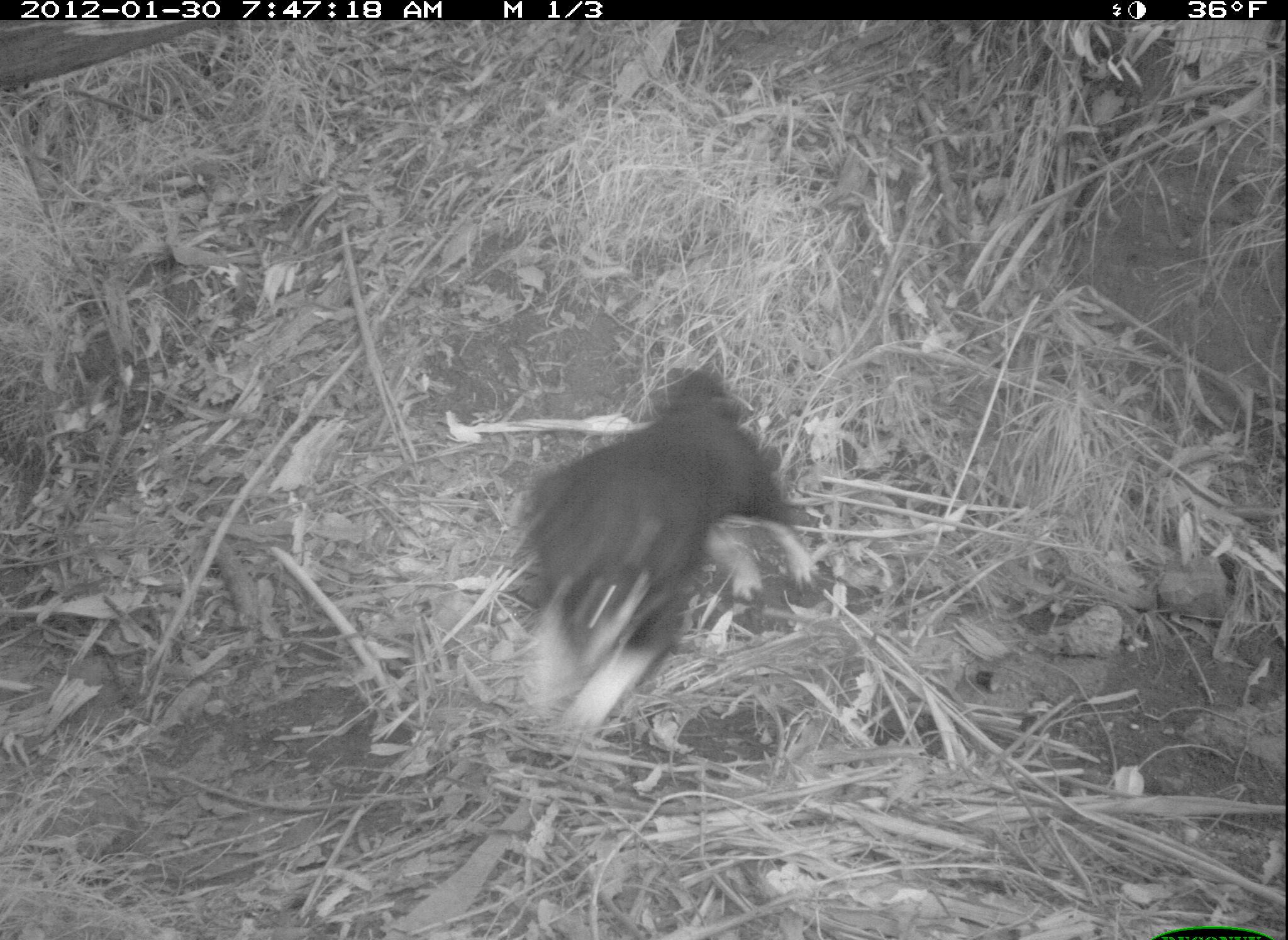} &
         \includegraphics[width=0.5\textwidth,height=0.5\textwidth]{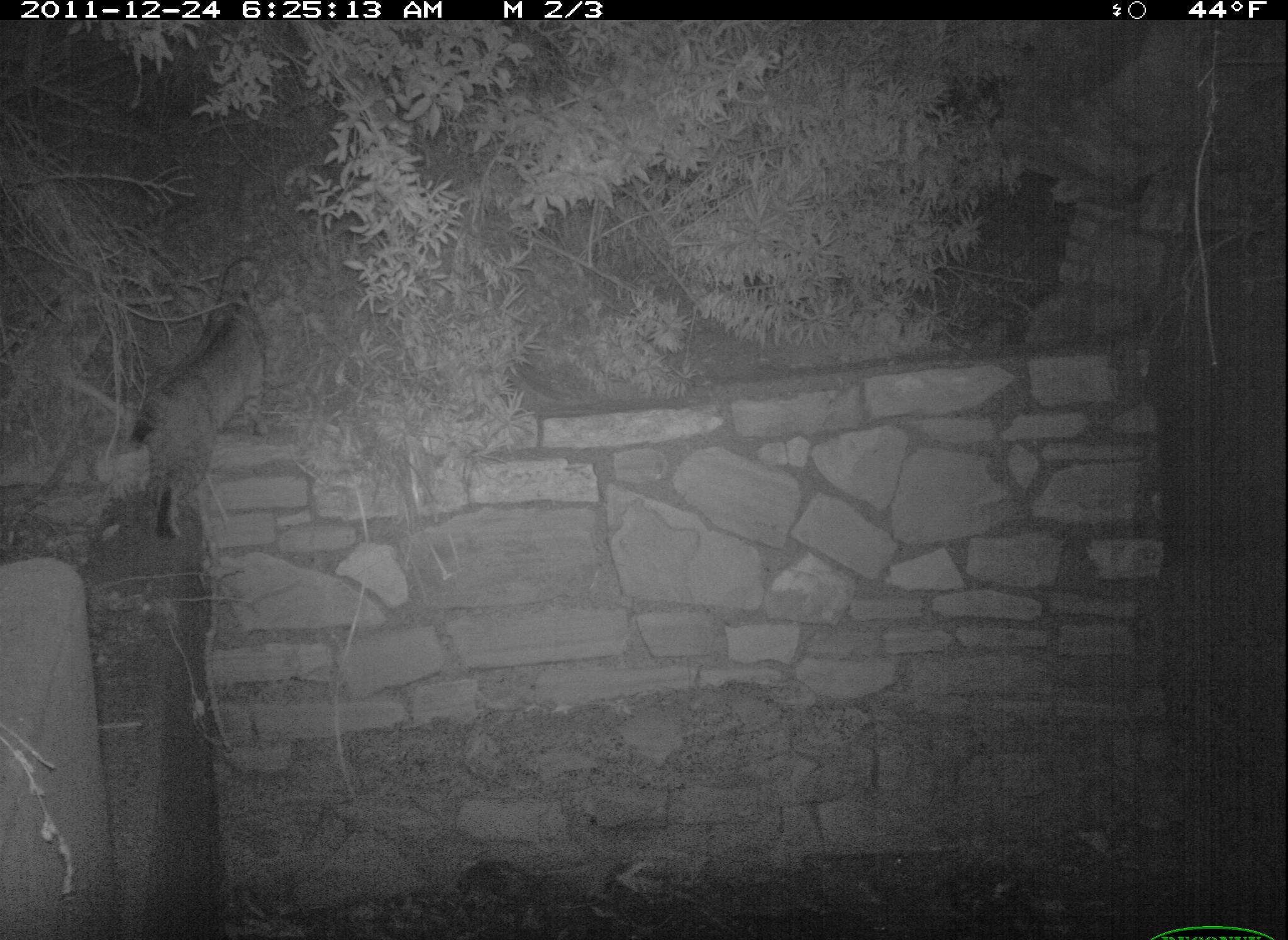} &
         \includegraphics[width=0.5\textwidth,height=0.5\textwidth]{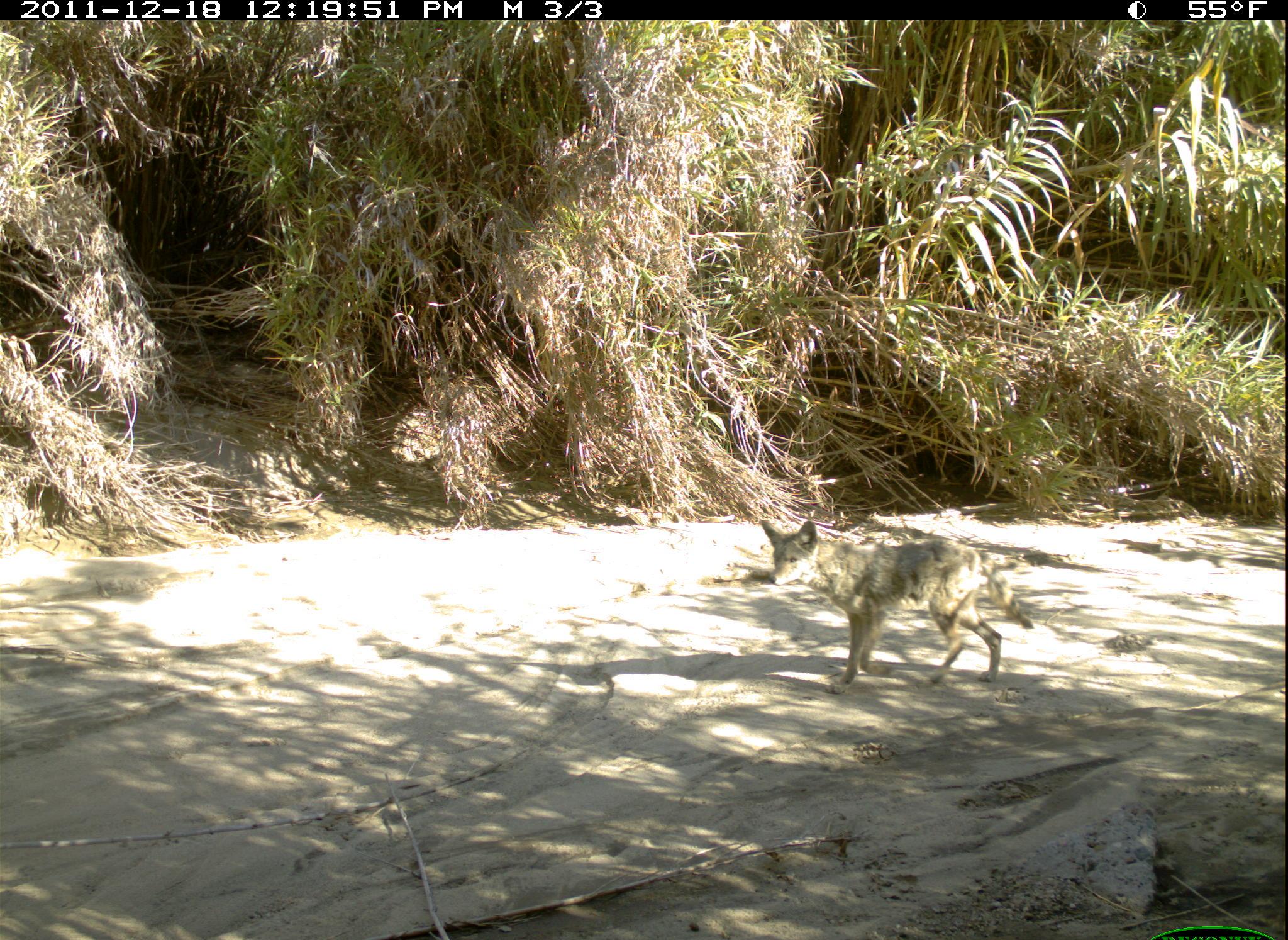} \\
         \includegraphics[width=0.5\textwidth,height=0.5\textwidth]{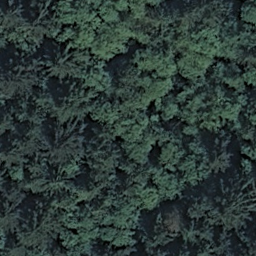} &
         \includegraphics[width=0.5\textwidth,height=0.5\textwidth]{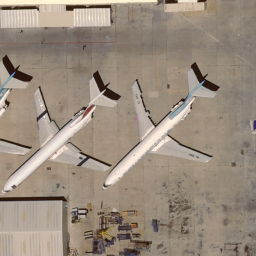} &
         \includegraphics[width=0.5\textwidth,height=0.5\textwidth]{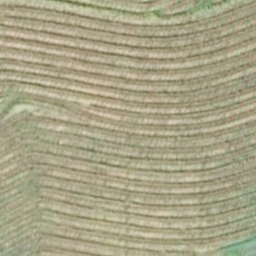} &
         \includegraphics[width=0.5\textwidth,height=0.5\textwidth]{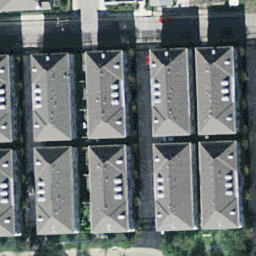} &
         \includegraphics[width=0.5\textwidth,height=0.5\textwidth]{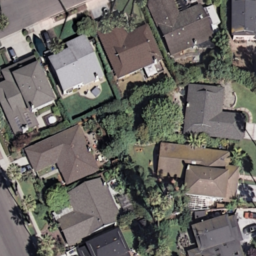} \\
         \includegraphics[width=0.5\textwidth,height=0.5\textwidth]{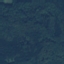} &
         \includegraphics[width=0.5\textwidth,height=0.5\textwidth]{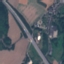} &
         \includegraphics[width=0.5\textwidth,height=0.5\textwidth]{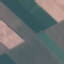} &
         \includegraphics[width=0.5\textwidth,height=0.5\textwidth]{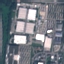} &
         \includegraphics[width=0.5\textwidth,height=0.5\textwidth]{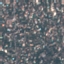} \\
         \includegraphics[width=0.5\textwidth,height=0.5\textwidth]{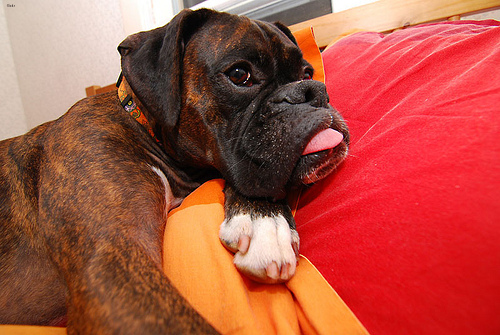} &
         \includegraphics[width=0.5\textwidth,height=0.5\textwidth]{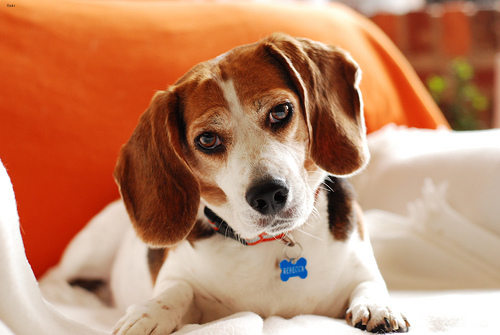} &
         \includegraphics[width=0.5\textwidth,height=0.5\textwidth]{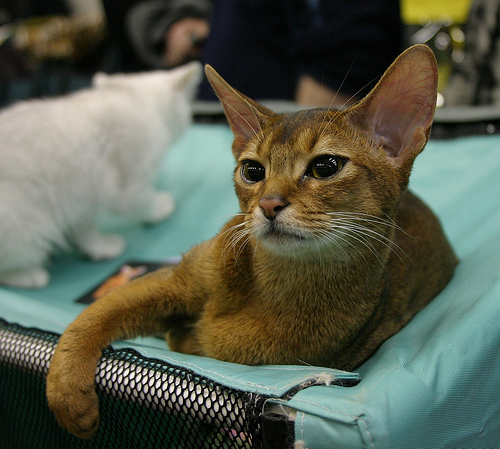} &
         \includegraphics[width=0.5\textwidth,height=0.5\textwidth]{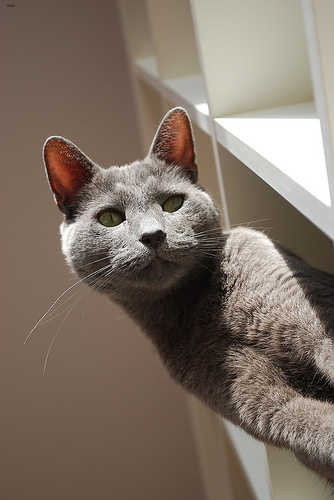} &
         \includegraphics[width=0.5\textwidth,height=0.5\textwidth]{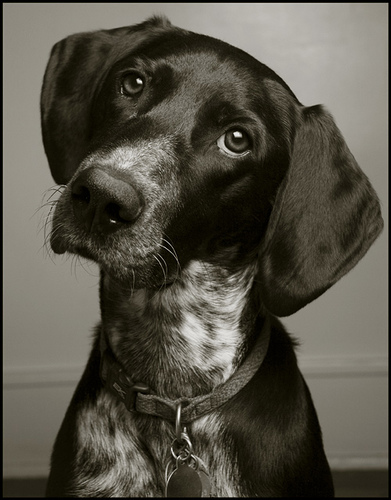} \\
         \includegraphics[width=0.5\textwidth,height=0.5\textwidth]{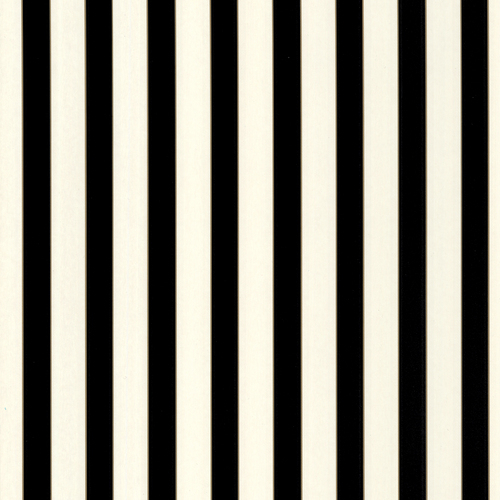} &
         \includegraphics[width=0.5\textwidth,height=0.5\textwidth]{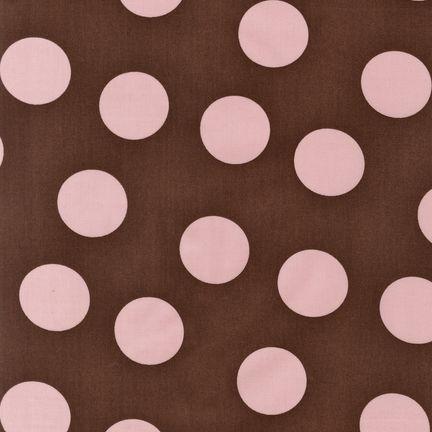} &
         \includegraphics[width=0.5\textwidth,height=0.5\textwidth]{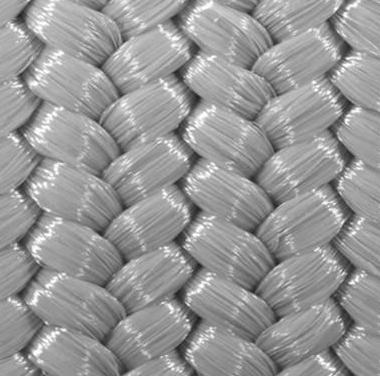} &
         \includegraphics[width=0.5\textwidth,height=0.5\textwidth]{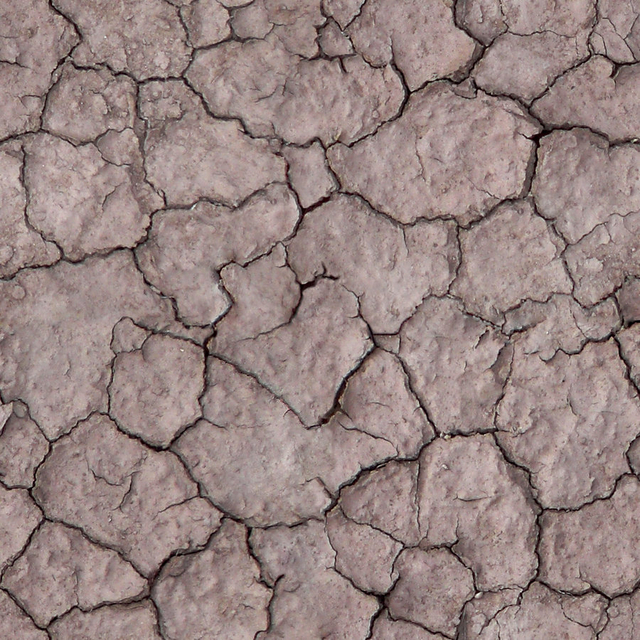} &
         \includegraphics[width=0.5\textwidth,height=0.5\textwidth]{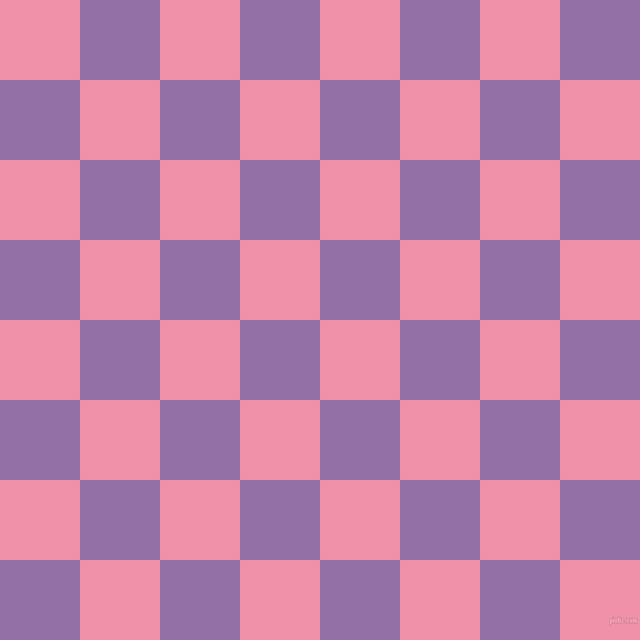} \\
    \end{tabular}}
    \caption{Examples of images in each dataset considered. From top to bottom: \texttt{HAM10000}, \texttt{FIVES}, \texttt{CheXpert}, \texttt{TerraIncognita}, \texttt{UCMerced}, \texttt{EuroSAT}, \texttt{Oxfort Pets}, \texttt{DTD}. Images are reshaped to square size for a more compact visualization.\\}
    \label{fig:datasets}
\end{figure}

The use of ICL in the field of computer vision is a relatively novel approach in research, and with limited studies exploring it in depth, we decided to use the same datasets as in \citep{yixing2024manyshot}. At the time of writing, \citep{yixing2024manyshot} is the only work that provides sufficient information to serve as a benchmark for our solution.
The selected datasets, summarized in Table \ref{tab:datasets}, span several domains (natural imagery, medical imagery, pattern imagery, remote sensing) and tasks (multi-class, multi-label and fine-grained classification). Some sample images for each dataset are reported in Figure \ref{fig:datasets}. 
For a fair comparison, we use the same subsets used in \citep{yixing2024manyshot}: the demo set is used to populate the knowledge base, enabling the system to retrieve tailored examples for each query input, while the test set is used to evaluate the solution.
In the scaling experiments, we increase the number of demonstrating examples retrieved by the retriever without ensuring class balance.
Although it is impossible to guarantee that the datasets were not used in the training of the MLLM, the baseline column in Table \ref{tab:accuracycomparison}, representing the model zero-shot accuracy, provides insights into whether these are familiar contexts. For example, it is evident that \texttt{FIVES} and \texttt{CheXpert} are unfamiliar domains, as the accuracy on these datasets is comparable to that of a random guess classifier.

\begin{table*}[h]
\caption{Summary of the datasets used in the experiments. For each dataset we report the name of the dataset with its reference, a short discretion of the content of the dataset, the classification task performed, the size of the demo and test set, and the number of classes.}
\label{tab:datasets}
\centering
\begin{tabular}{lllcc} 
\toprule
\textbf{Dataset} & \textbf{Description} & \textbf{Task}  & \textbf{Demo/Test size}  & \textbf{Num. of classes} \\
\midrule
\texttt{HAM10000} \citep{ham10000dataset} & skin disease & single-label classification & 805 / 210 & 7 \\
\texttt{FIVES} \citep{fivesdataset} & eye disease & single-label classification & 400 / 120 & 4 \\
\texttt{CheXpert} \citep{cheexpertdataset} & lung disease & multi-label classification & 200 / 150 & 5 \\
\texttt{TerraIncognita} \citep{terraincognitadataset} & animal species & single-label classification & 1035 / 270 & 9 \\
\texttt{UCMerced} \citep{ucmerceddataset} & land use  & single-label classification & 1470 / 420 & 21 \\
\texttt{EuroSAT} \citep{eurosatdataset} & land cover & single-label classification & 1000 / 300 & 10 \\
\texttt{Oxford Pets} \citep{oxfordpetsdatasets} & pet & fine-grained single-label classification & 1750 / 700 & 35 \\
\texttt{DTD} \citep{cimpoi2013describingtextureswild}  & texture & single-label classification & 2350 / 940 & 47 \\
\bottomrule
\end{tabular}
\end{table*}


\subsection{Settings}
The costs associated with using Gemini's APIs are, at the time of writing, high enough to pose a significant budgetary constraint.
As a result, we opted to test our solution on a wide variety of datasets to better explore its generalization capabilities, rather than focusing on achieving peak performance with the proposed approach.

The experiments are conducted using the same number of demonstrating examples for each instance in the test set of every dataset. Specifically, for each dataset, experiments are repeated using 5, 10 and 50 examples, maintaining the same RAG configuration (Embeddings, Retrieval system, and Generator model).

Finally, for the smallest datasets (i.e., \texttt{FIVES} and \texttt{CheXpert}), which entail reduced costs, additional experiments are also conducted using 100 examples.

\subsection{Evaluation}
The proposed system is evaluated using the following metrics to permit a direct comparison with the performance reported in \citep{yixing2024manyshot}:
\begin{itemize}
  \item[-] Accuracy: for all single-label classification task datasets;
  \item[-] Macro-averaged F1: for the multi-label classification task dataset \texttt{CheXpert}.  
\end{itemize}
This approach highlights the improvements introduced by the RAG in the context of ICL while avoiding the need to rerun all the baseline tests provided by \citep{yixing2024manyshot}.
To ensure the validity of the comparison, a subset of baseline tests was performed in the initial phase.


\section{Results}

We compare the performance of the proposed Visual RAG with two different approaches: the first one acts as a baseline, which is the MLLM model queried without providing any examples in context (i.e., zero-shot classification); the second one is Many-shot ICL, i.e., the approach used by \citep{yixing2024manyshot}, which constitutes the state of the art: 
the approach randomly selects from the knowledge base the examples to be provided in context to query the MLLM.
The best classification accuracy obtained by the three different approaches across all the 8 datasets considered is reported in Table~\ref{tab:accuracycomparison}.

\begin{table*}[h]
\caption{Best accuracy obtained in all the experiments with different approaches. Baseline: accuracy achieved without providing any examples to the MLLM model (i.e., zero-shot, also corresponding to the generator-only ablation study). Many-shot ICL: accuracy achieved by \cite{yixing2024manyshot}, which provides randomly selected examples from the Knowledge Base (KB).  VisualRAG: accuracy obtained using the proposed VisualRAG solution. Retriever-only: accuracy achieved by the retriever-only ablation study. As a further reference, in the last column we also report the accuracy reached by state-of-the-art supervised methods.}
\label{tab:accuracycomparison}
\centering
\begin{tabular}{lccccc} 
\toprule
 & \multicolumn{4}{c}{\textbf{Accuracy} $(\uparrow$)} \\
\textbf{Dataset} & \textbf{Baseline (zero-shot)}  & {\textbf{Many-shot ICL}} \citep{yixing2024manyshot}  & \textbf{Visual RAG} & \textbf{Ablation: Retriever-only} & \textbf{Supervised SOTA} \\
\midrule
\texttt{HAM10000} & 33.33 & 56.46 (+23.13) & \bf{56.67} (+23.34) & 44.29 (+10.96) & 96.49 \citep{lan2022fixcaps}\\
\texttt{FIVES}  & 25.83 & 55.00 (+29.17) & \bf{64.17} (+38.34) & 45.83 (+20.00) & 74.40 \cite{silva2025foundation}\\
\texttt{CheXpert}  & 22.16 & 42.23 (+20.08) & \bf{42.43} (+20.27) & - & 50.44 \citep{nguyen2022learning}\\
\texttt{TerraIncognita} & 59.63 & \bf{66.67} (+7.04) & 66.42 (+6.79) & 41.04 (-18.59) & 69.60 \citep{zhang2023towards}\\
\texttt{UCMerced}  & 91.19  & 98.57 (+7.38) & \bf{99.29} (+8.10) & 86.51 (-4.68) & 100.00 \citep{gesmundo2022continual}\\
\texttt{EuroSAT}  & 36.24 & 74.16 (+37.92) & \bf{83.22} (+46.98) & 73.78 (+37.54) & 99.24 \citep{wang2024mtp}\\
\texttt{Oxford Pets} & 85.29 & \bf{97.43} (+12.14) & 97.14 (+11.85) & 80.40 (-4.89) & 97.40 \citep{fini2024multimodal}\\
\texttt{DTD}  & 69.89 & 83.19 (+7.50) & \bf{84.26} (+14.37) & 61.70 (-7.95) & 90.00 \citep{ortiz2024task} \\
\bottomrule
Average  & 52.95 & 71.71 (+18.76) & \bf{74.20} (+21.25) & 61.94 (+8.99) & 84.70\\
\end{tabular}
\end{table*}

These results 
highlight how the Visual RAG improved the accuracy of the MLLM used, even outperforming Many-shot ICL \citep{yixing2024manyshot} in almost all datasets, showing an excellent ability to work in very different contexts.
Notably, the only datasets where Visual RAG did not achieve the performance levels of Many-shot ICL are \texttt{TerraIncognita} and \texttt{OxfordPets}. However, in these datasets the accuracy gap is very low: 0.25\% for \texttt{TerraIncognita} and 0.29\% for \texttt{OxfordPets}.
 
In all the other six datasets, and particularly in \texttt{FIVES}, \texttt{EUROSAT} and \texttt{DTD}, it is evident that the MLLM significantly benefits from the tailored selection of examples, achieving a marked improvement in performance compared to the Many-shot ICL solution. This behavior could be attributed to the reduction of noise (i.e., images irrelevant to the specific query) within the context, as previously observed and highlighted for textual RAGs \citep{barnett2024sevenfailurepointsengineering} \citep{liu2023lostmiddlelanguagemodels}. 
\texttt{FIVES} serves as one of the most representative examples of the capabilities of the proposed solution: the baseline accuracy of 25.83\%, which is almost equivalent to that of a random classifier (i.e., 25.00\% considering that the dataset has 4 classes), suggests that the MLLM lacks prior knowledge of this visual domain. Even in these conditions our solution is able to obtain +38.34\% in accuracy (+9.17\% compared to Many-shot ICL), showing how the solution is able to enhance MLLM knowledge even when the context is completely new and cannot be traced back to knowledge already possessed. As an additional reference, Table \ref{tab:accuracycomparison} also reports the best results reached by supervised methods specifically designed for each dataset, although using different data splits. From these results we can observe how the average performance of Visual RAG is at more than half-way (approx. 67\%) between zero-shot and the best supervised methods.


\begin{figure*}
\centering
\includegraphics[width=\textwidth]{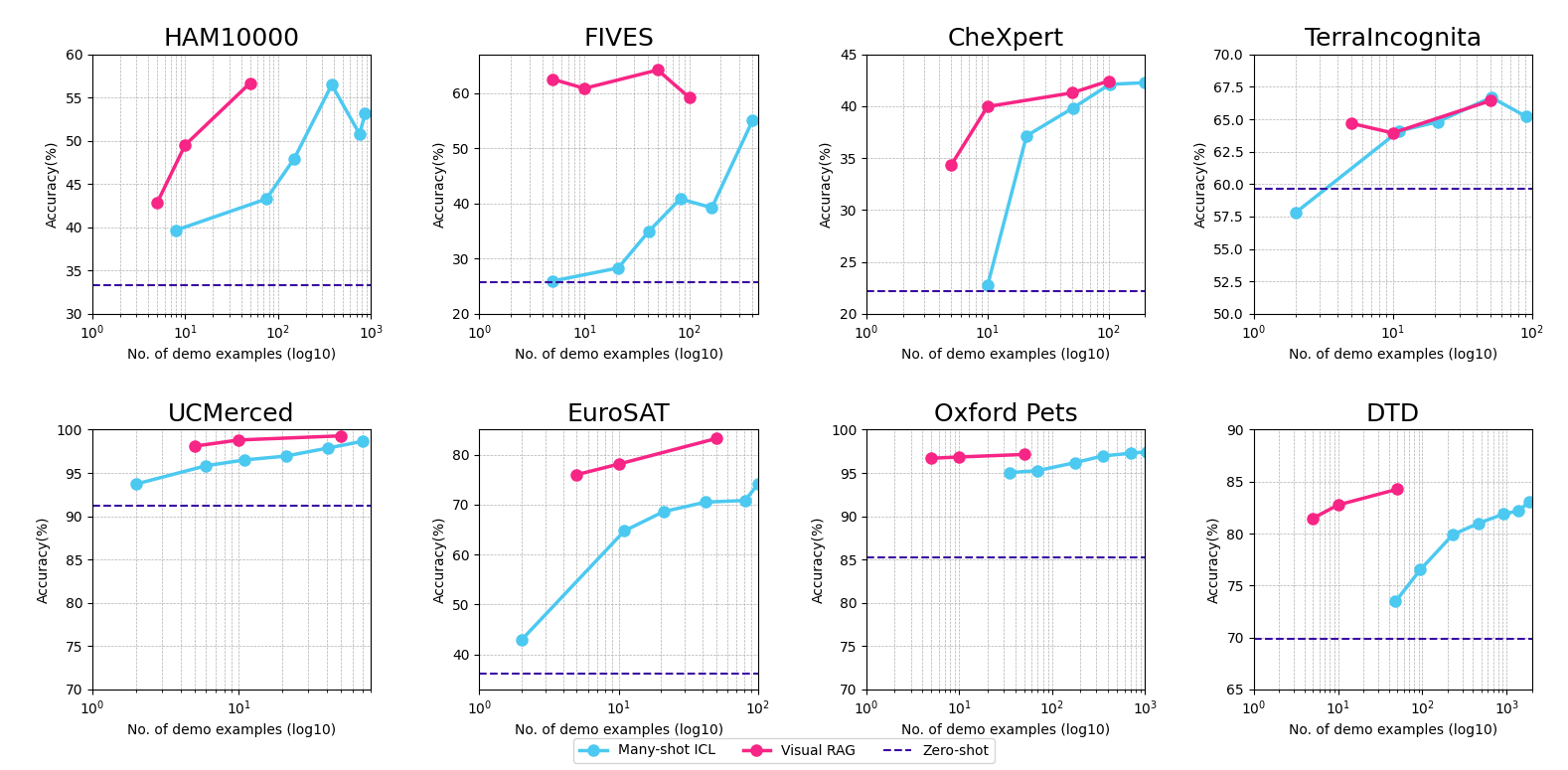}
\caption{Accuracy comparison between the proposed Visual RAG solution, Many-shot ICL \citep{yixing2024manyshot}, and zero-shot baseline on the eight datasets considered.\\}
\label{fig:accuracyplot}
\end{figure*}

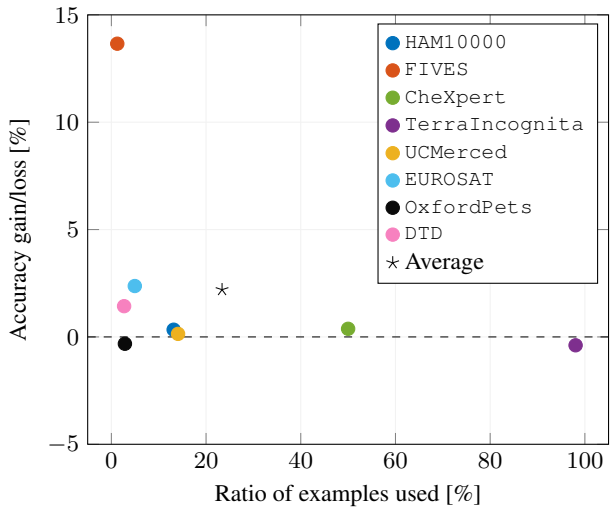
\begin{figure}
\centering
\begin{tikzpicture} 
\begin{axis}[
		xmin=-5,
		xmax=105,
		xlabel = {Ratio of examples used [\%]},
		ylabel = {Accuracy gain/loss [\%]},
            ymax=15,
            ymin=-5,
            grid=both,
            grid style={black!5},
            legend style={legend cell align=left},
        ]
\addplot [only marks, color1, mark size=2.5pt] coordinates
{(13.157895,                 0.336402) };
\addplot [only marks, color2, mark size=2.5pt] coordinates
{(1.250000,                13.657029) };
\addplot [only marks, color5, mark size=2.5pt] coordinates
{(50.000000,                 0.378519) };
\addplot [only marks, color4, mark size=2.5pt] coordinates
{(98.039216,                -0.389922) };
\addplot [only marks, color3, mark size=2.5pt] coordinates
{(14.084507,                 0.141887) };
\addplot [only marks, color6, mark size=2.5pt] coordinates
{(4.950495,                 2.370690) };
\addplot [only marks, color7, mark size=2.5pt] coordinates
{(2.860412,                -0.318112) };
\addplot [only marks, color8, mark size=2.5pt] coordinates
{(2.691066,                 1.432527) };
\addplot [only marks, color7, mark size=2.5pt, mark=star] coordinates
{(23.379198759523845,                 2.201127441445733) };
\addplot [color7, dashed] coordinates
{(-5,0) (105,0) };
\legend{\footnotesize{\texttt{HAM10000}}, \footnotesize{\texttt{FIVES}}, \footnotesize{\texttt{CheXpert}}, \footnotesize{\texttt{TerraIncognita}}, \footnotesize{\texttt{UCMerced}}, \footnotesize{\texttt{EUROSAT}}, \footnotesize{\texttt{OxfordPets}}, \footnotesize{\texttt{DTD}}, Average}
\end{axis}
\end{tikzpicture}
\caption{Efficiency comparison between the proposed Visual RAG solution and Many-shot ICL \citep{yixing2024manyshot}. For each dataset, a comparison is made between the experiment with the highest accuracy obtained using Many-shot ICL, and the experiment with the lowest number of examples that was able to match or surpass that accuracy using Visual RAG. In cases where the accuracy was not improved, the experiment exhibiting the closest performance was selected. The plot reports the accuracy gain of Visual RAG with respect to Many-shot ICL, versus the ratio of demo samples used.\\ \\}
\label{fig:efficiencyplot}
\end{figure}

The complete performance of the compared solutions for the different number of demonstrating examples considered are reported in Figure~\ref{fig:accuracyplot}. The plots reported hint at a further key benefit of our method:
 a significant improvement in terms of efficiency. Visual RAG demonstrates the capability to achieve comparable or even superior accuracy to the state-of-the-art method Many-shot ICL \citep{yixing2024manyshot}, while requiring a substantially reduced number of demonstrating examples. 
In order to further highlight this aspect, in Figure \ref{fig:efficiencyplot} we plot the difference in the best accuracy between Visual RAG and Many-shot ICL versus the ratio of examples used by Visual RAG with respect to those of Many-shot ICL to obtain the best accuracy. 


On average, Visual RAG uses only 23\% of the demonstrating example typically required by Many-shot ICL. Moreover, excluding \texttt{CheXpert} and \texttt{TerraIncognita} where the retriever encounters difficulties due to specific dataset characteristics (which will be examined later), our approach required an astonishingly low average of 6.5\% demonstrating examples.
In the case of \texttt{OxfordPets}, \texttt{DTD} and \texttt{FIVES} this quantity is further reduced to less than 3\%.

This confirms that our solution, compared to Many-shot ICL \citep{yixing2024manyshot}, is much more efficient in terms of data usage.
Greater efficiency implies reduced input size (i.e., number of tokens), leading to lower computational costs and processing time.

Finally, our solution demonstrates the same remarkable characteristic observed in \citep{yixing2024manyshot}, where increasing the use of demonstrating examples directly improves the achieved accuracy. This highlights the scalability and effectiveness of the approach.
But, as expected, as we can observe in \texttt{FIVES} and \texttt{CheXpert} datasets, when the number of examples provided approaches the total number of examples in the demo set, the performance begins to converge towards that obtained with Many-shot ICL. 

In contrast, the datasets where our solution encountered some difficulties are: 
\begin{itemize}
  \item[-] \texttt{OxfordPets}: This dataset, unlike the others, challenges the model's fine-grained classification capabilities due to the high inter-class similarity. Even the results presented in \citep{yixing2024manyshot} indicate that the MLLM benefits less from the examples provided in the context compared to other datasets. This could be attributed to the fact that the embeddings currently used by the model to analyze the input images are not sufficiently detailed to capture the subtle differences necessary to achieve the highest accuracy.  
  Despite these challenges, the gap between our solution and the best accuracy with Many-shot ICL remains minimal, even when using only less than 3\% of the examples. Moreover, the graph still highlights an upward trend, suggesting that the difference in accuracy could potentially be bridged through further experiments retrieving a larger number of demonstrating examples.
  \item[-] \texttt{TerraIncognita}: This dataset is composed of images that are not very focused on the subject to be classified. In fact, being photographs taken by camera traps, they often have the subject appearing in a very small part of the image, often out of focus or with artifacts due to subject movement. All these characteristics put a strain on the capabilities of the retriever, who often provides images where the similarity is more related to the background than to the subject of the image. The difficulties of the retriever module act as a bottleneck of our solution to the point that the results obtained are the same as those of the Many-shot ICL solution. Future research could refine the retriever to allow it to better focus on the main subject of the images, solving/mitigating the problem that emerged here.
  \item[-] \texttt{CheXpert}: Unlike other datasets, this one specifically tests the ability to perform multi-label classification. We have decided to include this dataset in this list because although the results were overall improved, the retriever encountered some challenges that are worth analyzing. Specifically, similar to \texttt{TerraIncognita}, the retriever's mechanism, which works on the overall similarity of the images, struggles to focus on specific elements. In a multi-label context, this characteristic compromises the quality of the images provided as examples: the examples retrieved are usually linked to the predominant label in the query and in these examples, the remaining elements are neglected, even partially. Future research could refine the retriever to allow it focus on specific subjects of the images, solving/mitigating this problem.
\end{itemize}

\begin{figure*}
    \includegraphics[width=\textwidth]{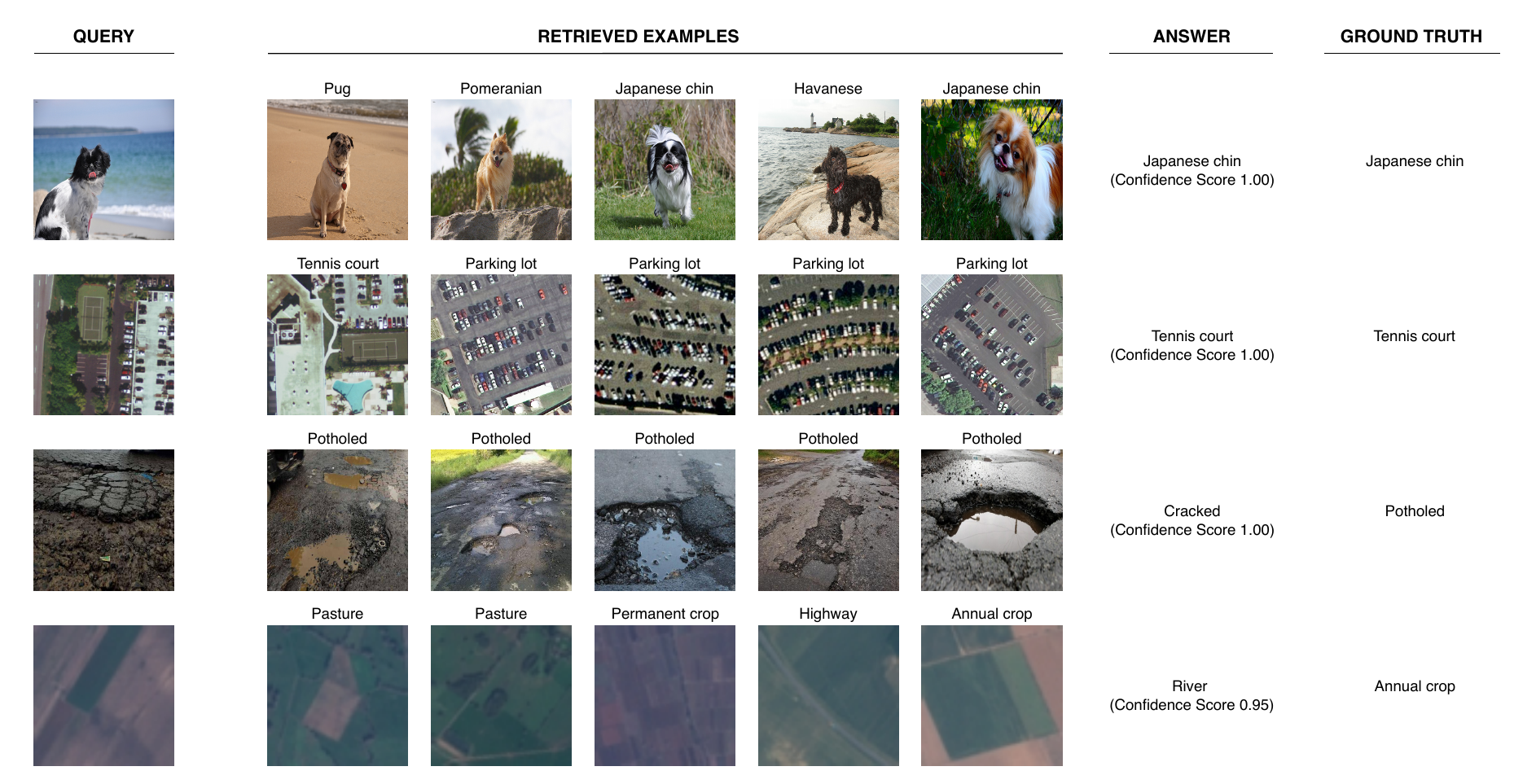}
    \caption{Example results of the proposed Visual RAG in the experimental configuration where five demonstration examples are provided by the retriever. Each row represents a different case. For each case we report: the query image, the five retrieved examples, the answer provided by Visual RAG with the corresponding confidence score, and the ground truth label. From top to bottom the query images belong to: \texttt{Oxford Pets}, \texttt{UCMerced}, \texttt{DTD}, and \texttt{EuroSAT} dataset.\\}
    \label{fig:results}
\end{figure*}

Some example results of the proposed Visual RAG are reported in Figure \ref{fig:results}. The results are relative to the case where five demonstration examples are provided. For each example, we provide: the query image, the selected demonstrating examples with the corresponding labels, the answer provided by Visual RAG, and the ground truth label.
The first example comes from the \texttt{Oxford Pets} dataset. The query image represents a "Japanese chin" on the beach. We can see how the retriever returns two examples containing the same dog breed and three examples containing the same background. This is reasonable since the embedding model does not know what the task will be. The generator instead is able to solve this ambiguity and returns the correct answer, understanding that the focus is on the dog breed.
The second example comes from the \texttt{UCMerced} dataset. The query image depicts a tennis court surrounded by two parking lots, both being possible labels. As for the previous example, the retriever returns examples belonging to both categories (i.e., "tennis court" and "parking lot"). Also in this case, the generator solves the ambiguity providing the correct label, understanding that in this dataset if a tennis court is completely visible in the image, it has to be labeled as such.
The third example comes from the \texttt{DTD} dataset, with the query image depicting a "potholed" texture. Although the retriever returns only "potholed" examples, the generator, relying on its previous knowledge, predicts the image as belonging to the "cracked" class. Looking at the other images in the dataset labeled as "cracked", we can see a high similarity with the query image. 
The fourth example comes from the \texttt{EuroSAT} dataset, with the query image depicting an "annual crop". Although the retriever returns also examples of the ground-truth class, the generator predicts the "river" class.

\section{Ablation}
In this section we conduct ablation studies on the key components of the Visual RAG in all the experiments.
\begin{itemize}
  \item[-] Generator-only ablation study: the accuracy reported in Table \ref{tab:accuracycomparison} in the Baseline column shows the performance achieved without using the retriever module. This column represents the capacity of the system using only the generator model (MLLM), without providing any demonstrating examples (i.e., zero-shot). The results indicate that our solution enhances the generator classification capabilities by leveraging the demonstrating examples in all the datasets. We can see the improvement in accuracy (the difference in accuracy between our solution and the baseline) inside the brackets in the column Visual RAG, showing an average increase of 21.25\%, ranging from 6.79\% in \texttt{TerraIncognita} to 46.98\% in \texttt{EuroSAT}.
  \item[-] Retriever-only ablation study: the accuracy reported in Table \ref{tab:accuracycomparison} in the column Retriever-only shows the performance achieved without using the generator. This ablation study is conducted to assess the classification capabilities of a system composed solely of the retriever. In this system, the label of the query is simply predicted as the most frequent class present among the examples provided. In the computation of the accuracy, in the case of a tie between the correct and other class(es), the prediction is considered incorrect. 
\end{itemize}
The ablation studies conducted reveal that the combined approach offers a significant advantage over using either component alone: by leveraging the strengths of both the retriever and the generator, we are able to achieve superior accuracy across all the scenarios considered.
The most notable improvements are observed in cases where the generator has limited or no contextual knowledge like \texttt{HAM10000}, \texttt{FIVES} and \texttt{EUROSAT}. The retriever effectively compensates for these limitations, providing the necessary context to enhance the generator's decision-making.
However, the benefits of our solution extend beyond these specific cases. Even when the generator has a strong understanding of the context, our combined approach still delivers enhanced performance.
These flexibility and robustness make our solution a compelling choice for a wide range of applications.

\section{Conclusion}

In-Context Learning (ICL) holds a significant potential in computer vision applications, with the primary challenge being to identify the method that leverages it most effectively. 
The proposed Visual RAG method has demonstrated superior performance compared to the state of the art.
The key properties emerging from Visual RAG are:
\begin{itemize}
  \item[-] Expansion of the use cases: It enables a straightforward expansion of knowledge regarding image classification in MLLMs, allowing to leverage the models capabilities in novel contexts they are unfamiliar with.
  \item[-] Accuracy improvement: Demonstrating examples guide the model, even in familiar domains, resulting in more accurate predictions. Furthermore, providing only carefully selected examples using a retriever ensures that only relevant information for the task is included. This noise reduction leads to improved accuracy compared to state-of-the-art approaches \citep{yixing2024manyshot}, even when using the same Knowledge Base (KB).
  \item[-] Efficiency: By including only query-relevant context information, our approach achieves an average reduction of about 77\% in the number of demonstrating examples required compared to state-of-the-art methods. The smaller context size accelerates inference time, significantly reduces associated costs, and alleviates the problem of narrow context windows in MLLMs.
  \item[-] Scalability: Our results, consistent with those reported in \citep{yixing2024manyshot}, show that even when using a retriever module, increasing the number of demonstrating examples continues to lead to improved performance.
  \item[-] Rapid updates: in case of new examples available, or any type of KB update, our solution allows a rapid adaptation without costs and with immediate effect.
\end{itemize}

While our solution has shown significant improvements, our research has identified several refinements that could further enhance performance, as for example alternative retrieval methodologies, different index configurations, and prompts. 
However, as our ablative studies have shown, the overall performance is highly correlated with the capabilities of its constituent components. Consequently, even without these additional refinements, our solution is expected to benefit significantly from advancements in the underlying embedding, retrieval and MLLM models.

As future works we plan to explore alternative retrieval methodologies and index configurations, such as hierarchical or semantic-aware indexing, to improve the precision of retrieved examples, particularly for datasets with fine-grained or multi-label classification tasks.
We also plan to investigate the use of advanced embedding models or fine-tuned embeddings tailored to specific visual domains to address challenges like inter-class similarity or subject-background separation. We will also extend Visual RAG to other computer vision tasks to evaluate its versatility beyond image classification.




\bibliography{mybibfile}


\newpage
\appendix

\section{Prompt used in the Visual RAG solution}
In this section we report the used in the proposed Visual RAG solution, which corresponds to the prompt used for image classification experiments in \cite{yixing2024manyshot}. 
\begin{verbatim}
prompt = ""
for demo in demo_examples:

prompt += f"""<<IMG>>Given the image above, answer the following question-
using the specified format.

Question: What is in the image above?
Choices: {str(class_desp)}
Answer Choice: {demo.answer}
"""

prompt += f"""<<IMG>>Given the image above, answer the following question-
using the specified format.

Question: What is in the image above?
Choices: {str(class_desp)}
Please respond with the following format:
---BEGIN FORMAT TEMPLATE---
Answer Choice: [Your Answer Choice Here]
Confidence Score: [Your Numerical Prediction Confidence Score Here From 0 To 1]
---END FORMAT TEMPLATE---
Do not deviate from the above format. Repeat the format template for the answer."""
\end{verbatim}

\end{document}